\documentclass[10pt,twocolumn,letterpaper]{article}

\usepackage{cvpr}      %

\usepackage{url}

\usepackage{amsmath}
\usepackage{booktabs}       %
\usepackage{amsfonts}       %
\usepackage{nicefrac}       %
\usepackage{microtype}      %
\usepackage{xcolor}         %
\usepackage{graphicx}
\usepackage{multirow}
\usepackage{subcaption}
\usepackage{wrapfig}
\usepackage{algorithm}
\usepackage{algorithmic}
\usepackage{graphicx}
\usepackage{amssymb}
\usepackage{array}
\usepackage{natbib}
\usepackage{bbm}

\definecolor{cvprblue}{rgb}{0.21,0.49,0.74}
\definecolor{gray}{gray}{0.95}
\definecolor{green}{HTML}{009000} 
\definecolor{red}{HTML}{ea4335} 
\usepackage[pagebackref,breaklinks,colorlinks,allcolors=cvprblue]{hyperref}

\def\paperID{18555} %
\def\confName{CVPR}
\def\confYear{2025}

\title{Emphasizing Discriminative Features for Dataset Distillation in Complex Scenarios}

\author{
    Kai Wang$^{1, 2*}$, Zekai Li$^{1, 2*}$, Zhi-Qi Cheng$^{2\dag}$, Samir Khaki$^{3}$, Ahmad Sajedi$^{3}$, Ramakrishna\\ \ Vedantam$^{4}$, Konstantinos N Plataniotis$^{3}$, Alexander Hauptmann$^{2}$, Yang You$^{1\dag}$
    \\[0.04cm]
    $^{1}$NUS,
    $^{2}$CMU,
    $^{3}$University of Toronto,
    $^{4}$Independent Researcher
}

\newlength\savewidth\newcommand\shline{\noalign{\global\savewidth\arrayrulewidth
  \global\arrayrulewidth 1pt}\hline\noalign{\global\arrayrulewidth\savewidth}}

\def\floor#1{\lfloor #1 \rfloor}
\renewcommand{\paragraph}[1]{\vspace{1mm}\noindent\textbf{#1}}

\newcommand{\tablestyle}[2]{\setlength{\tabcolsep}{#1}\renewcommand{\arraystretch}{#2}\centering\footnotesize}

\newcommand{\error}[1]{$_{\pm#1}$}
\begin{document}
\maketitle

\renewcommand{\thefootnote}{}
\footnotetext[0]{$^{*}$equal contribution (kai.wang@comp.nus.edu.sg, lizekai@u.nus.edu)\ \ $^{\dag}$advising contribution. Part of this work was completed at CMU.}

\begin{abstract}
Dataset distillation has demonstrated strong performance on simple datasets like CIFAR, MNIST, and TinyImageNet but struggles to achieve similar results in more complex scenarios.
In this paper, we propose \textbf{EDF} (\textbf{e}mphasizes the \textbf{d}iscriminative \textbf{f}eatures), a dataset distillation method that enhances key discriminative regions in synthetic images using Grad-CAM activation maps.
Our approach is inspired by a key observation: in simple datasets, high-activation areas typically occupy most of the image, whereas in complex scenarios, the size of these areas is much smaller.
Unlike previous methods that treat all pixels equally when synthesizing images, EDF uses Grad-CAM activation maps to enhance high-activation areas.
From a supervision perspective, we downplay supervision signals produced by lower trajectory-matching losses, as they contain common patterns.
Additionally, to help the DD community better explore complex scenarios, we build the Complex Dataset Distillation (Comp-DD) benchmark by meticulously selecting sixteen subsets, eight easy and eight hard, from ImageNet-1K.
In particular, EDF consistently outperforms SOTA results in complex scenarios, such as ImageNet-1K subsets.
Hopefully, more researchers will be inspired and encouraged to improve the practicality and efficacy of DD. 
Our code and benchmark have been made public at \href{https://github.com/NUS-HPC-AI-Lab/EDF}{NUS-HPC-AI-Lab/EDF}.
\end{abstract}

\section{Introduction}
\label{sec:intro}
Dataset Distillation (DD) has been making remarkable progress since it was first proposed by~\cite{wang2020dataset}.
Currently, the mainstream of DD is matching-based methods~\citep{DC, DSA, DM, Cazenavette2022DatasetDB}, which first extract patterns from the real dataset, then define different types of supervision to inject extracted patterns into the synthetic data.
On several simple benchmarks, such as CIFAR~\citep{Krizhevsky2009LearningML} and TinyImageNet~\citep{Le2015TinyIV}, existing 
DD methods can achieve lossless performance~\citep{guo2024lossless, li2024prioritizealignmentdatasetdistillation}. 

\begin{figure}[t]
    \centering
    \begin{subfigure}{0.17\textwidth}
        \centering
        \includegraphics[width=\textwidth]{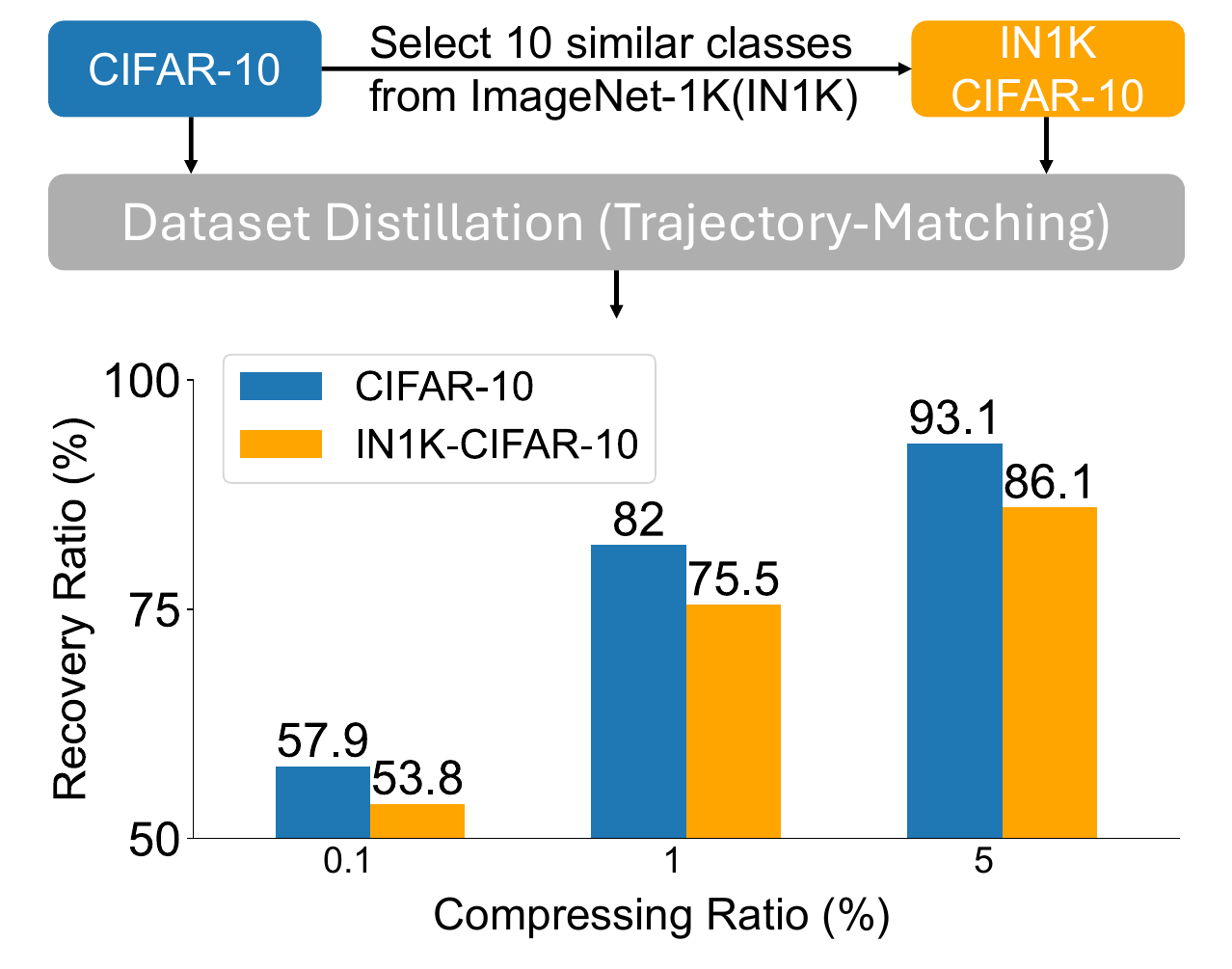}
        \caption{The performance of dataset distillation drops remarkably in complex scenarios.}
        \label{cifar_vs_in1k}
    \end{subfigure}
    \hfill
    \begin{subfigure}{0.3\textwidth}
        \centering
        \includegraphics[width=\textwidth]{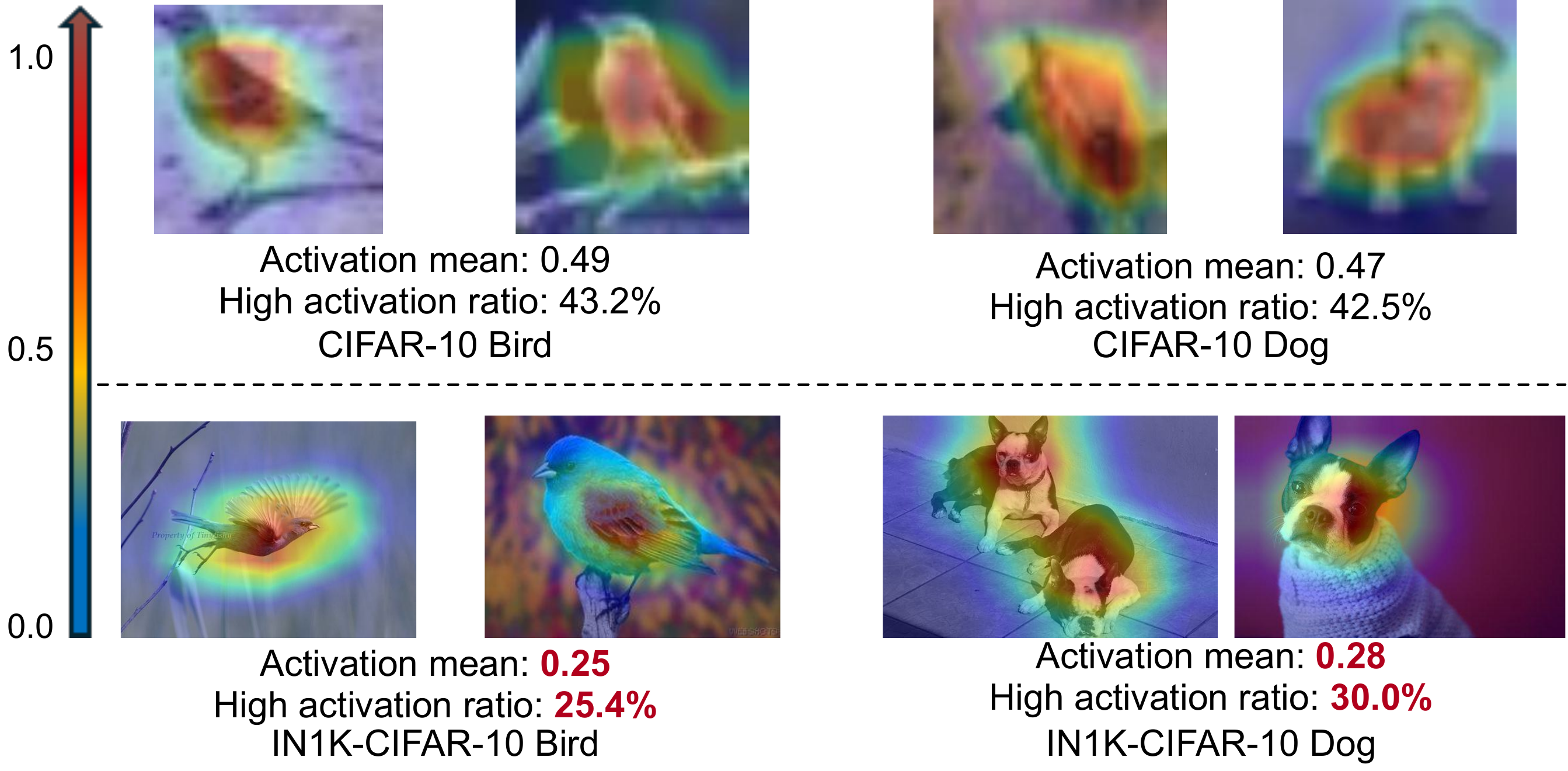}
        \caption{Images from IN1K-CIFAR-10 have much lower activation means and smaller highly activated areas.}
        \label{cam_comp}
    \end{subfigure}
    \caption{\textbf{(a)} DD recovery ratio (distilled data accuracy over full data accuracy) comparison between CIFAR-10 and IN1K-CIFAR-10. We use trajectory matching for demonstration. \textbf{(b)} Comparison between Grad-CAM activation map statistics of CIFAR-10 and IN1K-CIFAR-10. The ratio refers to the percentage of pixels whose activation values are higher than 0.5.}
    \label{fig1}
\end{figure}

However, DD still struggles to be practically used in real-world applications, \textit{i.e.}, images in complex scenarios are characterized by significant variations in object sizes and the presence of a large amount of class-irrelevant information.
To show that current DD methods fail to achieve satisfying performance in complex scenarios, we apply trajectory matching~\citep{guo2024lossless} on a 10-class subset from ImageNet-1K extracted by selecting similar classes of CIFAR-10, called IN1K-CIFAR-10.
As depicted in Figure~\ref{cifar_vs_in1k}, under three compressing ratios, DD's performances\footnote[1]{We compare recovery ratios because datasets are different.} on IN1K-CIFAR-10 are consistently worse than those on CIFAR-10.

To figure out the reason, we take a closer look at the ImageNet-1K and CIFAR-10 from the data perspective.
One key observation is that the percentage of discriminative features in the complex scenario, which can be visualized by Grad-CAM~\citep{Selvaraju2016GradCAMVE}, is much lower.
From Figure~\ref{cam_comp}, CIFAR-10 images are sticker-like, and activation maps have higher means and larger highly activated areas.
By contrast, activation maps of the IN1K-CIFAR-10 subset exhibit much lower activation means and smaller highly activated areas.
Previous methods~\citep{Cazenavette2022DatasetDB, guo2024lossless} treat all pixels of synthetic images \textbf{equally}. 
Therefore, when applying these methods to more complex scenarios, the large ratio of low-activation areas leads to non-discriminative features dominating the learning process, resulting in a performance drop.

\begin{figure}[t]
    \centering
    \begin{subfigure}{0.47\textwidth}
        \centering
        \includegraphics[width=\textwidth]{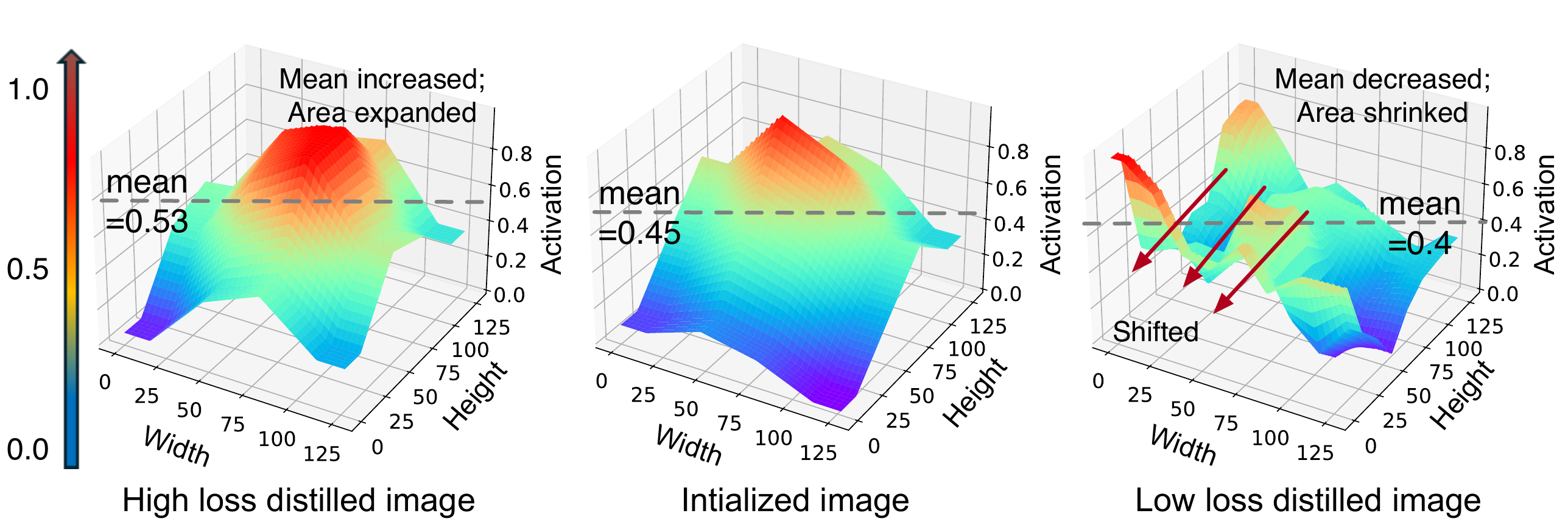}
        \caption{High-loss supervision increases activation means and expands the high-activation area, while low-loss supervision reduces the activation mean and shifts to the wrong discriminative area.}
        \label{fig2a}
    \end{subfigure}
    \hfill
    \begin{subfigure}{0.47\textwidth}
        \centering
        \includegraphics[width=\textwidth]{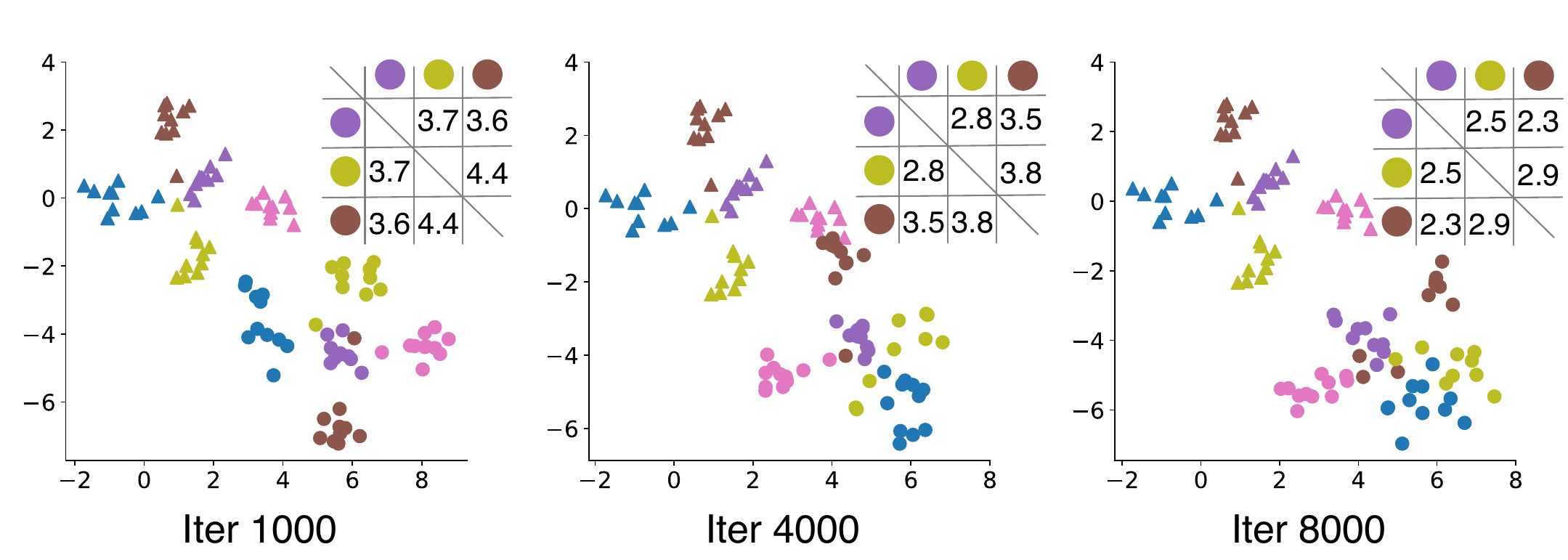}
        \caption{Triangles and circles represent real and synthetic image features, respectively. As the distillation with low-loss only proceeds, more and more common patterns are introduced to synthetic images.}
        \label{fig2b}
    \end{subfigure}
    \caption{\textbf{(a)} Grad-CAM activation maps of the image with initialization, high-loss supervision distillation, and low-loss supervision distillation. \textbf{(b)} t-SNE visualization of image features with \textbf{only} low-loss supervision. Different colors represent different classes. The top right is \textbf{inter-class distance} computed by the average of point-wise distances.}
    \label{fig2}
\end{figure}

As for supervision, taking trajectory matching as an example, we investigate the impact on synthetic images of supervision signals brought by different trajectory-matching losses.
Specifically, we compare the performance between i) using only supervision from low trajectory-matching losses to update synthetic images and ii) using only supervision from high trajectory-matching losses to update synthetic images.
The effect on a single image is shown in Figure~\ref{fig2a}. 
Low-loss supervision reduces the mean of Grad-CAM activation maps and shrinks the high-activation area (also shifted).
By contrast, high-loss supervision increases the activation mean and expands the high-activation region.

Additionally, we visualize the inter-class feature distribution for a broader view.
In Figure~\ref{fig2b}, we show the t-SNE of features of synthetic images distilled by only low trajectory-matching losses.
As the distillation proceeds, synthetic image features of different classes continuously come closer, and the confusion among classes becomes more severe, which is likely caused by common patterns. 
The above two phenomenons confirm that low-loss supervision primarily reduces the representation of discriminative features and embeds more common patterns into synthetic images, harming DD's performance.

Based on the above observations, we propose to \textbf{Emphasize Discriminative Features} (EDF), built on trajectory matching.
To synthesize more discriminative features in the distilled data, we enable discriminative areas to receive more updates compared with non-discriminative ones.
This is achieved by guiding the optimization of synthetic images with gradient weights computed from Grad-CAM activation maps.
Highly activated pixels are assigned higher gradients for enhancement.
To mitigate the negative impact of common patterns, the EDF distinguishes between different supervision signals by dropping those with a low trajectory matching loss according to a drop ratio.

To help the community explore DD in complex scenarios, we extract new subsets from ImageNet-1K with various levels of complexity and build the Complex DD benchmark (Comp-DD).
The complexity levels of these new subsets are determined by the average ratios of high-activation areas (Grad-CAM activation value $>$ predefined threshold).
We run EDF and several typical DD methods on partial Comp-DD and will release the full benchmark for future studies to further improve performance.

In summary, EDF consistently achieves state-of-the-art (SOTA) performance across various datasets, underscoring its effectiveness.
On some ImageNet-1K subsets, EDF achieves lossless performance. 
To the best of our knowledge, we are the first to achieve lossless performance on ImageNet-1K subsets.
We build the Complex Dataset Distillation benchmark based on complexity, providing convenience for future research to continue improving DD's performance in complex scenarios.

\section{Method}

\begin{figure*}[t]
    \centering
    \includegraphics[width=0.9\textwidth]{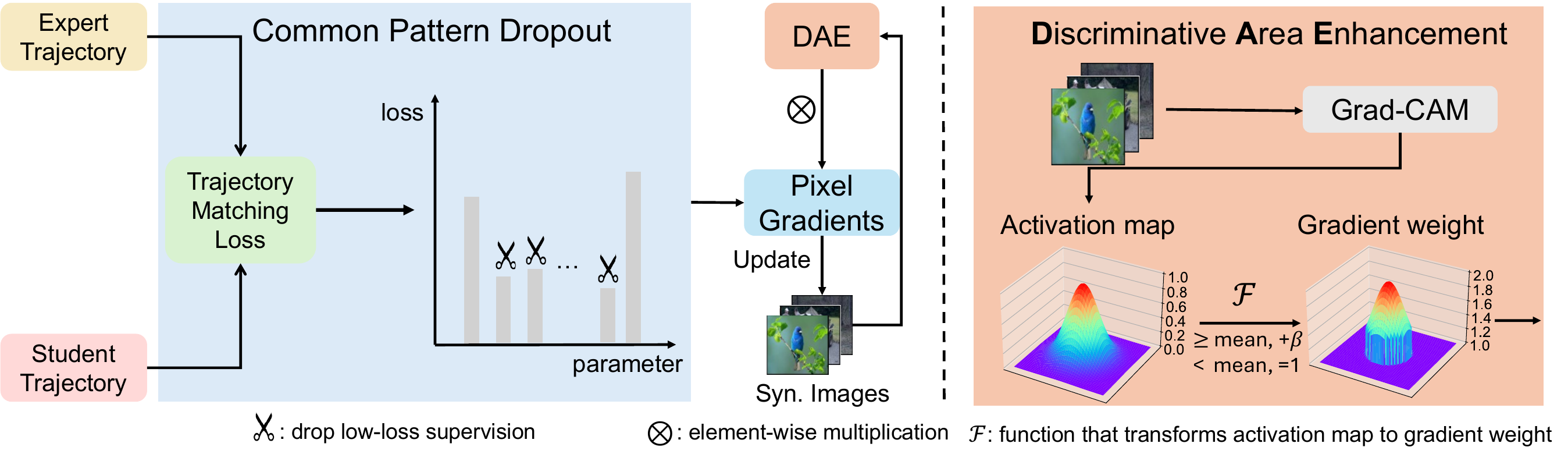}
    \caption{Workflow of \textbf{Emphasize Discriminative Features} (EDF). EDF comprises two modules:~(1)~\textit{Common Pattern Dropout}, which filters out low-loss signals, and the~(2)~\textit{Discriminative Area Enhancement}, which amplifies gradients in critical regions. $\beta$ denotes the enhancement factor. "mean" denotes the mean activation value of the activation map.}
    \label{fig3}
\end{figure*}

Our approach, \textbf{Emphasize Discriminative Features} (EDF), enhances discriminative features of synthetic images. As shown in Figure~\ref{fig3}, EDF first trains trajectories on real images $T$ and synthetic images $S$ and computes the trajectory matching loss. Then, \textit{Common Pattern Dropout} filters out low-loss supervision signals, retaining high-loss ones for backpropagation. After obtaining gradients for the synthetic images, \textit{Discriminative Area Enhancement} uses dynamically extracted Grad-CAM activation maps to rescale pixel gradients, focusing updates on discriminative regions.
Although EDF is built on the trajectory-matching backbone, the two modules can be easily applied to other matching-based methods with minor modifications.

\subsection{Common Pattern Dropout}
This module reduces common patterns in supervision by matching expert and student trajectories on real and synthetic data, then removing low-loss elements. This ensures only meaningful supervision enhances the model's ability to capture discriminative features.

\paragraph{Trajectory Generation and Loss Computation.}
To generate expert and student trajectories, we first train agent models on real data for $E$ epochs, saving the resulting parameters as expert trajectories, denoted by $\{\theta_{t}\}_{0}^{E}$. At each distillation iteration, we randomly select an initial point $\theta_{t}$ and a target point $\theta_{t+M}$ from these expert trajectories.
Similarly, student trajectories are produced by initializing an agent model at $\theta_{t}$ and training it on the synthetic dataset, yielding the parameters $\{\hat{\theta}_{t}\}_{0}^{N}$. The trajectory matching loss is the distance between final student parameters $\hat{\theta}_{t+N}$ and expert’s target parameters $\theta_{t+M}$, normalized by the initial difference:
\begin{equation}
    L = \frac{||\hat{\theta}_{t+N} - \theta_{t+M}||^2}{||\theta_{t+M} - \theta_t||^2}.
\end{equation}
Instead of directly summing this loss, we decompose it into an array of individual losses between corresponding parameters in the expert and student trajectories, represented as $L = \{l_1, l_2, \dots, l_{P}\}$, where $P$ is the number of parameters, and $l_i$ is the loss associated with the $i$-th parameter.

\paragraph{Low-loss Element Dropping.}
Our analyses of Figure~\ref{fig2a} and~\ref{fig2b} show that signals of low-loss trajectories typically introduce common patterns, which hinder the learning of key discriminative features, particularly in complex scenarios. 
To address this, we sort the array of losses computed from the previous step in ascending order. 
Using a predefined dropout ratio $\alpha$, we discard the smallest $\floor{\alpha \cdot P}$ losses ($\floor{}$ denotes the floor function), which are assumed to capture common, non-discriminative features. The remaining losses are summed and normalized to form the final supervision:
\begin{equation}
    L \ \xrightarrow{\text{sort}} \ 
    L^{'} = \{\textcolor[HTML]{9F9F9F}{\underbrace{l_1, l_2,\cdots,l_{\floor{\alpha \cdot P}}}_{\rm dropout\,},}\underbrace{l_{\floor{\alpha \cdot P} + 1},\cdots, l_{P}}_{\rm sum\& normalize}\},
\end{equation}
where $L^{'}$ represents the updated loss array after dropping the lowest $\floor{\alpha \cdot P}$ elements. The remaining losses, $l_{\floor{\alpha \cdot P} + 1}, \dots, l_P$, are then summed and normalized to form the final supervision signal.

\subsection{Discriminative Area Enhancement}
After the pruned loss from \textit{Common Pattern Dropout} is backpropagated, this module amplifies the importance of discriminative regions in synthetic images. Grad-CAM activation maps are dynamically extracted from the synthetic data to highlight areas most relevant for classification. These activation maps are then used to rescale the pixel gradients, applying a weighted update that prioritizes highly activated regions, thereby focusing the learning process on key discriminative features.

\paragraph{Activation Map Extraction.}
Grad-CAM generates class-specific activation maps by leveraging the gradients that flow into the final convolutional layer, highlighting key areas relevant to predicting a target class. To compute these maps, we first train a convolutional model $\mathcal{G}$ on the real dataset. Following the Grad-CAM formulation (Equation~\ref{eq3}), we calculate the activation map for each class $c$: $M^c \in \mathbb{R}^{IPC \times H \times W}$ on the synthetic images (IPC is the number of images per class). The activation map $M^c$ is a gradient-weighted sum of feature maps across all convolutional layers:
\begin{equation}
\label{eq3}
    \alpha^{c} = \frac{1}{Z}\sum_{h}\sum_{w} \frac{\partial y^c}{\partial A^{l}_{h,w}}, \quad M^{c} = ReLU(\sum_{l} \alpha^{c}_{l}A^{l}).
\end{equation}
$\alpha^c$ represents the weight of activation corresponding to the $l$-th feature map, $A^l$, computed by gradients. Finally, we concatenate $M^c$ of all images in class $c$ and obtain $M \in \mathbb{R}^{|S| \times H \times W}$.
Note that the capability of the model used to extract activation maps does not affect the performance.
We provide ablation in the Appendix~\ref{app_cam_models}.

\paragraph{Discriminative Area Biased Update.}
A major limitation of previous DD algorithms~\citep{Cazenavette2022DatasetDB, Du2022MinimizingTA, guo2024lossless} on the complex scenario is that they treat each pixel \textbf{equally} and provide no guidance for the distillation process on which area of synthetic images should be emphasized.
Therefore, we propose to update synthetic images in a biased manner.
Instead of treating each pixel equally, we enhance the significance of discriminative areas by guiding the optimization with activation maps extracted in the previous step.
We define the discriminative area of a synthetic image as the percentage of pixels with activation values above the mean since synthetic images are dynamically changing (see Section~\ref{sec:ablation} for discussion).
Specifically, we process activation maps from the previous step with a function $\mathcal{F}(M, \beta)$ to create weights for pixel gradients as follows:
\begin{equation}
\label{eq4}
    \mathcal{F}(M^{i}_{h,w}, \beta) =
    \begin{cases}
      1 & \text{if } M^{i}_{h, w} < \bar{M^{i}}, \\
      \beta + M^{i}_{h, w} & \text{if } M^{i}_{h, w} \geq \bar{M^{i}}.
    \end{cases}
\end{equation}
$M^{i}_{h, w}$ denotes the activation value of the $i$ the image at coordinate $(h, w)$, and $\bar{M^i}$ denotes the mean activation of $M^i$.
$\beta \geq 1$ is called the \textit{enhancement factor}.
Then, we rescale the gradient matrix of synthetic images by multiplying it with the weight matrix element-wise:
\begin{equation}
    (\nabla D_{syn})_{edf} = \nabla D_{syn} \odot \mathcal{F}(M, \beta).
\end{equation}
We drag gradients of discriminative areas to a higher range so that they receive more updates.

\paragraph{Dynamic Update of Activation Maps.}
As synthetic images are optimized, high-activation regions shift over time. To capture these changes, we recompute the activation maps every $K$ iterations, focusing updates on the most relevant areas. The frequency $K$ is a tunable hyperparameter, adjusted based on the learning rate of the synthetic images (see Section~\ref{sec:ablation} for details). This ensures evolving discriminative areas are accurately captured. The complete algorithm is provided in the Appendix~\ref{sec:pseudo_code}.

\section{Complex Dataset Distillation Benchmark}

\begin{figure}[t]
    \centering
    \begin{subfigure}{0.19\textwidth}
        \centering
         \includegraphics[width=\textwidth]{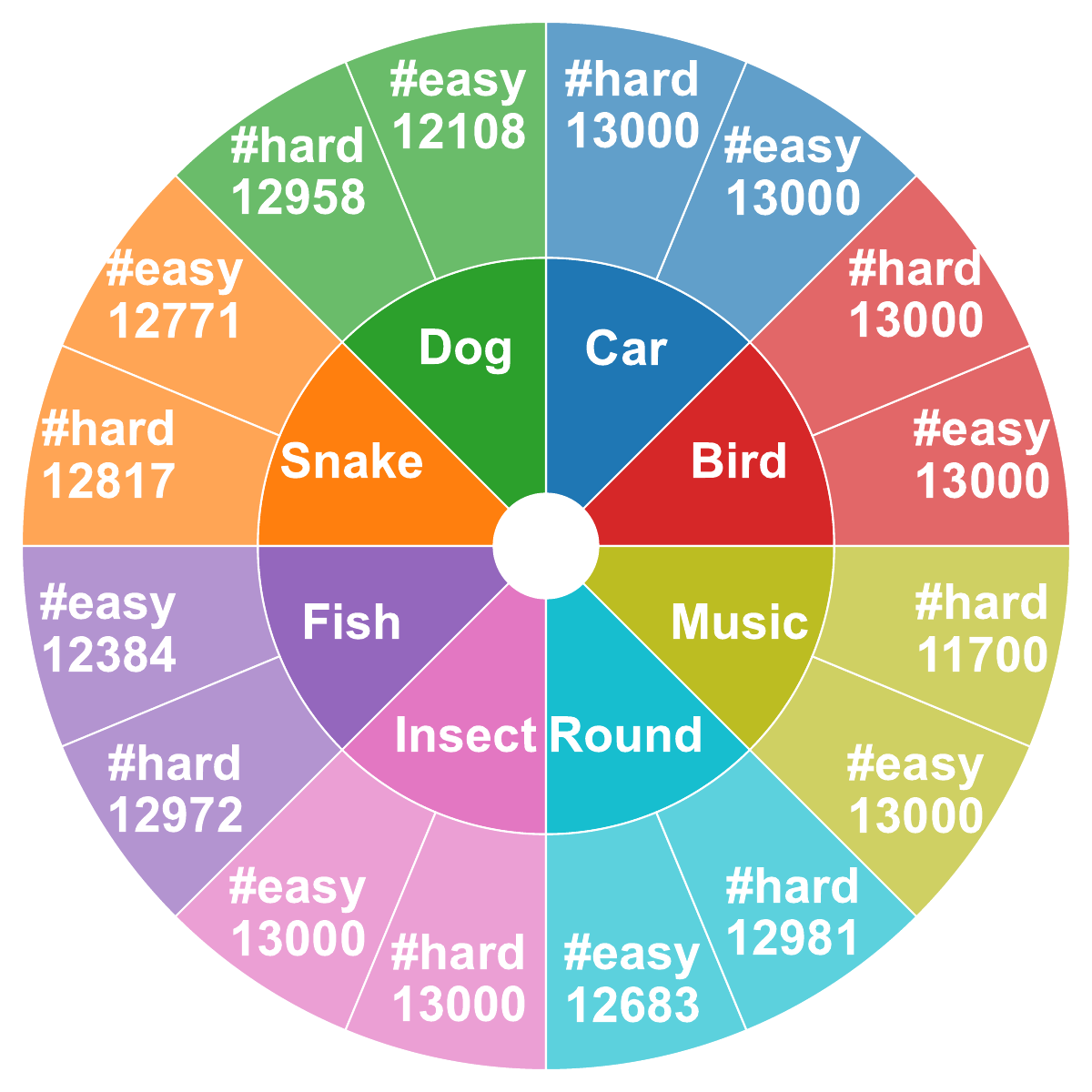}
         \caption{Number of images in each subset}
         \label{num_images_stat}
    \end{subfigure}
    \hfill
    \begin{subfigure}{0.28\textwidth}
        \centering
         \includegraphics[width=\textwidth]{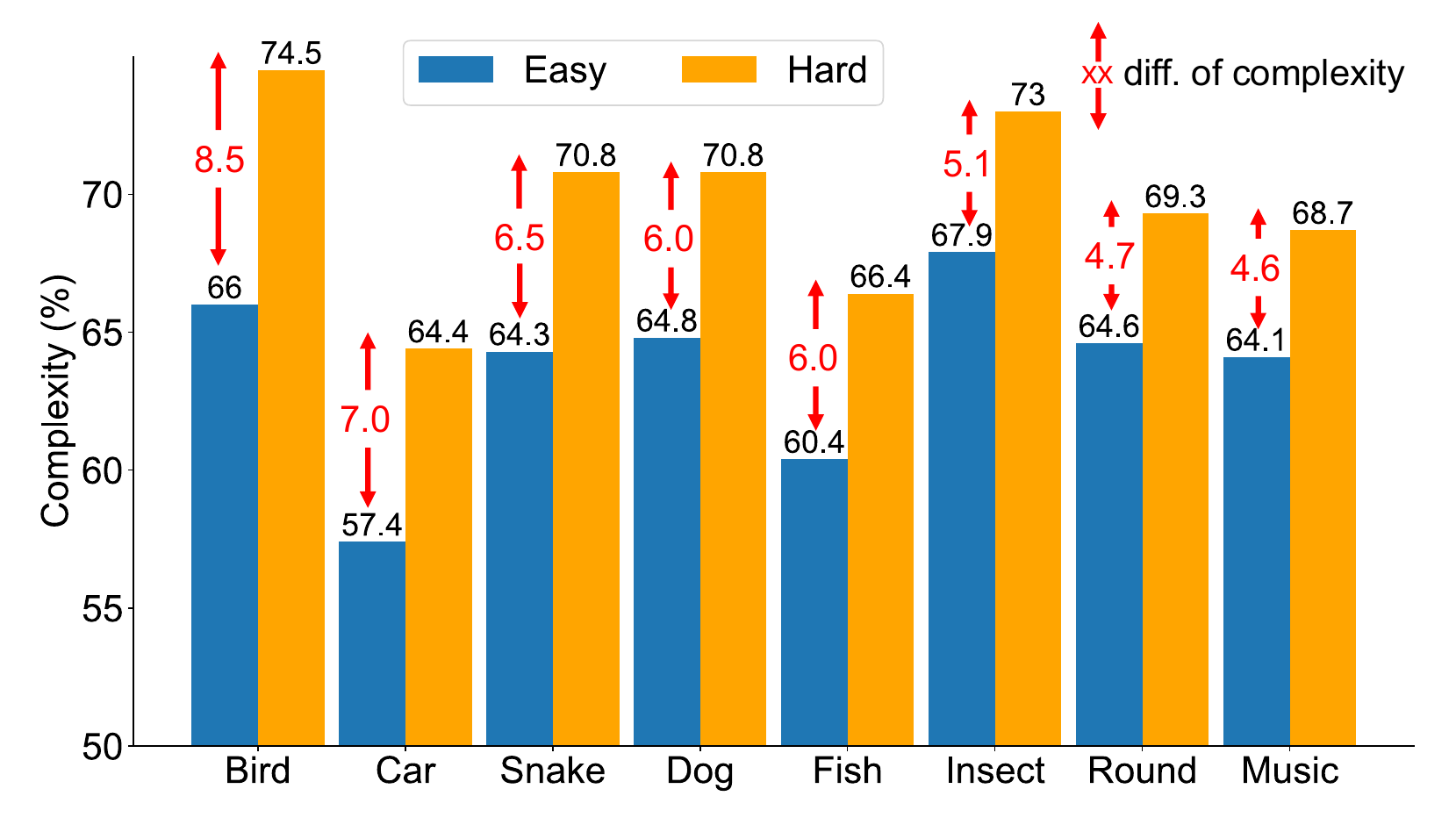}
         \caption{Complexity of Easy vs. Hard subsets in each category}
         \label{complexity_stat}
    \end{subfigure}
    \caption{\textbf{(a)} Statistics of the training set in the Comp-DD benchmark. Each subset contains 500 images in the validation set. \textbf{(b)} Comparison of subset-level complexity between easy and hard subsets across all categories. The complexity of hard subsets is higher than that of easy subsets.}
    \label{fig4}
\end{figure}

We introduce the \textbf{Comp}lex \textbf{D}ataset \textbf{D}istillation (Comp-DD) benchmark, which is constructed by selecting subsets from ImageNet-1K based on their complexity. 
This benchmark represents an early and pioneering effort to address dataset distillation in complex scenarios.
Although there are numerous benchmarks~\citep{Krizhevsky2009LearningML, Le2015TinyIV, cui2022dcbenchdatasetcondensationbenchmark} for simpler tasks, there is a notable absence of benchmarks designed specifically for complex scenarios. This gap presents a significant challenge to advancing research in this area and limits the practical application of dataset distillation.
To bridge this gap, we propose the first dataset distillation benchmark explicitly built around scenario complexity, aiming to promote further exploration within the DD community.

\paragraph{Complexity Metrics.}
We evaluate the complexity of an image by measuring the average size of high-activation regions of the Grad-CAM activation map. 
Using a pre-trained ResNet model, we first generate Grad-CAM activation maps for all images, class by class. 
For each image, we calculate the percentage of pixels with activation values above a predefined threshold $\alpha$ (set to 0.5 in our case), with higher percentages indicating lower complexity (more clarifications can be found in Appendix~\ref{app_comp_dd_metrics}). 
Formally, the complexity of the $i$-th image is computed as $1-\frac{\sum_{h}\sum_{w} \mathbbm{1}[M^{i}_{h,w} \geq \alpha]}{H \cdot W}$ where $\mathbbm{1}$ is the indicator function.
The complexity of each class is then determined by averaging the complexity scores across all images within that class.

\paragraph{Subset Selection.}
To reduce the influence of class differences, we select subsets from each \textit{category}, where a category consists of classes representing visually similar objects or animals of the same species. This approach allows us to focus on complexity while controlling for inter-class variability.
Specifically, we first manually identify representative categories in ImageNet-1K with sufficient numbers of classes ($\geq 20$). For each category, we rank the classes by complexity in descending order. Following established practice, we construct two ten-class subsets for each category: the \textit{easy} subset, comprising the ten least complex classes, and the \textit{hard} subset, comprising the ten most complex classes. The subset-level complexity is determined by averaging the complexity scores across all classes in each subset.

\paragraph{Statistics.}
We carefully selected eight categories from ImageNet-1K: Bird, Car, Dog, Fish, Snake, Insect, Round, and Music. Each category contains two ten-class subsets: one \textit{easy} and one \textit{hard}, with difficulty determined by the complexity metrics outlined above. Figure~\ref{num_images_stat} summarizes the number of training images in each subset, while all subsets contain 500 images in the validation set. To illustrate the difference between easy and hard subsets, Figure~\ref{complexity_stat} compares the subset-level complexity for each category. As expected, the hard subsets display significantly higher complexity than the easy subsets. For a detailed breakdown of the classes in each subset, please refer to Appendix~\ref{app_comp_dd_subset}.

\section{Experiments}
\subsection{Experimental Setup}
\vspace{-0.5em}
\paragraph{Datasets and Architecture.}
We conduct a comprehensive evaluation of EDF on six ten-class subsets~\citep{howard2019imagenette} of ImageNet-1K (ImageNette, ImageWoof, ImageMeow, ImageYellow, ImageFruit, and ImageSquawk) and a one-hundred-class subset (ImageNet100). 
Each subset contains ten classes, with approximately 13,000 images in the training set and 500 images in the validation set.
On the Comp-DD benchmark, we report the results of the Bird, Car, and Dog categories. 
For all experiments, we use a 5-layer ConvNet (ConvNetD5) as both the distillation and the evaluation architecture.
For cross-architecture evaluation (see results in Appendix~\ref{app_cross_arch}), we validate synthetic data accuracy on Alexnet~\citep{Krizhevsky2012ImageNetCW}, VGG11~\citep{Simonyan2014VeryDC}, ResNet18~\citep{He2015DeepRL}.

\begin{table*}[t]
\newcommand{\chart}{\includegraphics[height=2ex]{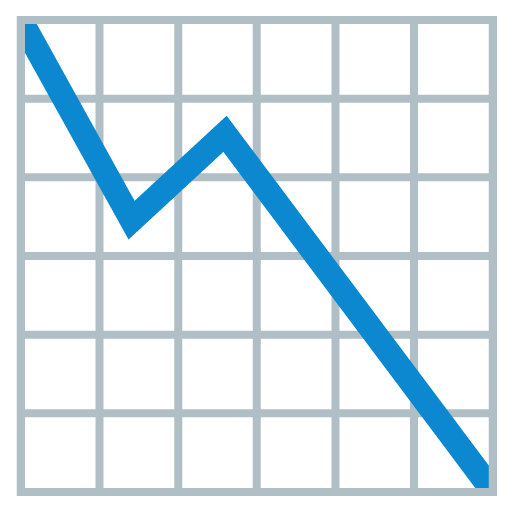}}
\centering
    \tablestyle{6pt}{1.1}
    
    \begin{tabular}{cc|ccccc|ccc|c}
    \multirow{2}{*}{Dataset}     & \multicolumn{1}{l|}{\multirow{2}{*}{IPC}} & \multicolumn{5}{c|}{DD} & \multicolumn{3}{c|}{Eval. w/ Knowledge Distillation}   & \multirow{2}{*}{Full}     \\
    & \multicolumn{1}{l|}{}                     & Random & MTT               & FTD        & DATM     & EDF      & SRe2L & RDED     & EDF &     \\ \shline
    \multirow{3}{*}{ImageNette}  & 1   & 12.6\error{1.5}    & 47.7\error{0.9}         & 52.2 \error{1.0} & 52.5\error{1.0} & \textbf{52.6\error{0.5}} & 20.8\error{0.2} & \textbf{28.9\error{0.1}} &  25.7\error{0.4}     & \multirow{3}{*}{87.8\error{1.0}} \\
                                 & 10  & 44.8\error{1.3}     & 63.0\error{1.3}          & 67.7\error{0.7}   & 68.9\error{0.8} & \textbf{71.0\error{0.8}} & 50.6\error{0.8} & 59.0\error{1.0} & \textbf{64.5\error{0.6}}  &                           \\
                                 & 50  & 60.4\error{1.4}     & \chart{}          & \chart{}   & 75.4\error{0.9} & \textbf{77.8\error{0.5}} & 73.8\error{0.6} & 83.1\error{0.6} & \textbf{84.8\error{0.5}}  &                           \\ \hline
    \multirow{3}{*}{ImageWoof}   & 1   & 11.4\error{1.3}     & 28.6\error{0.8} & 30.1\error{1.0}   & 30.4\error{0.7} & \textbf{30.8\error{1.0}} & 15.8\error{0.8} & 18.0\error{0.3}    & \textbf{19.2\error{0.2}}  & \multirow{3}{*}{66.5\error{1.3}} \\
                                 & 10  & 20.2\error{1.2}     & 35.8\error{1.8}          & 38.8\error{1.4}   & 40.5\error{0.6} & \textbf{41.8\error{0.2}} & 38.4\error{0.4} & 40.1\error{0.2} & \textbf{42.3\error{0.3}}  &                           \\
                                 & 50  & 28.2\error{0.9}     & \chart{}          & \chart{}   & 47.1\error{1.1} & \textbf{48.4\error{0.5}} & 49.2\error{0.4} & 60.8\error{0.5}    & \textbf{61.6\error{0.8}}  &                           \\ \hline
    \multirow{3}{*}{ImageMeow}   & 1   & 11.2\error{1.2}     & 30.7\error{1.6}          & 33.8\error{1.5}   & 34.0\error{0.5} & \textbf{34.5\error{0.2}} & \textbf{22.2\error{0.6}}      &  19.2\error{0.8}  & 20.8\error{0.5}  & \multirow{3}{*}{65.2\error{0.8}} \\
                                 & 10  &   22.4\error{0.8}   & 40.4\error{2.2}          & 43.3\error{0.6}   & 48.9\error{1.1} & \textbf{52.6\error{0.4}} & 27.4\error{0.5} & 44.2\error{0.6} &  \textbf{48.4\error{0.7}}       &                          \\
                                 & 50  & 38.0\error{0.5}     & \chart{}          & \chart{}   & 56.8\error{0.9} & \textbf{58.2\error{0.6}} & 35.8\error{0.7} & 55.0\error{0.6}  &  \textbf{58.2\error{0.9}}  & \\ \hline
    \multirow{3}{*}{ImageYellow} & 1   & 14.8\error{1.0}     & 45.2\error{0.8}          & 47.7\error{1.1}   & 48.5\error{0.4} & \textbf{49.4\error{0.5}} &  31.8\error{0.7}     & 30.6\error{0.2}  & \textbf{33.5\error{0.6}}  & \multirow{3}{*}{83.2\error{0.9}} \\
                                 & 10  & 41.8\error{1.1}     & 60.0\error{1.5}          & 62.8\error{1.4}   & 65.1\error{0.7} & \textbf{68.2\error{0.4}} &  48.2\error{0.5} & 59.2\error{0.5} & \textbf{60.8\error{0.5}}       &                           \\
                                 & 50  & 54.6\error{0.5}     & \chart{}          & \chart{}   & 70.2\error{0.8} & \textbf{73.2\error{0.8}} &  57.6\error{0.9} & 75.8\error{0.7} & \textbf{76.2\error{0.3}} & \\ \hline
    \multirow{3}{*}{ImageFruit}  & 1   & 12.4\error{0.9}     & 26.6\error{0.8}          & 29.1\error{0.9}   & 30.9\error{1.0} & \textbf{32.8\error{0.6}} &  23.4\error{0.5} & \textbf{33.8\error{0.4}} & 29.6\error{0.4}   & \multirow{3}{*}{64.4\error{0.8}} \\
                                 & 10  & 20.0\error{0.6}     & 40.3\error{0.5}         & 44.9\error{1.5}   & 45.5\error{0.9} & \textbf{46.2\error{0.6}} & 39.2\error{0.7} & 45.4\error{0.6} &  \textbf{48.4\error{0.8}}      &                           \\
                                 & 50  & 33.6\error{0.9}     & \chart{}          & \chart{}   & 50.2\error{0.5} & \textbf{53.2\error{0.5}} & 44.2\error{0.8} & 54.8\error{0.9} & \textbf{56.4\error{0.6}}  &                           \\ \hline
    \multirow{3}{*}{ImageSquawk} & 1   & 13.2\error{1.1}     & 39.4\error{1.5}          & 40.5\error{0.9}   & 41.1\error{0.6} & \textbf{41.8\error{0.5}} &  21.2\error{1.0}     & \textbf{33.8\error{0.6}}   & 30.5\error{0.5}   & \multirow{3}{*}{86.4\error{0.8}} \\
                                 & 10  & 29.6\error{1.5}     & 52.3\error{1.0}          & 58.4\error{1.5}   & 61.8\error{1.3} & \textbf{65.4\error{0.8}} &  39.2\error{0.3}     & 59.0\error{0.5} &  \textbf{59.4\error{0.6}}      &                           \\
                                 & 50  & 52.8\error{0.4}     & \chart{}          & \chart{}   & 71.0\error{1.2} & \textbf{74.8\error{1.2}} &  56.8\error{0.4} & 77.2\error{1.2} & \textbf{77.8\error{0.5}}  &                           \\ \hline
    \multirow{3}{*}{ImageNet100} & 1   &  2.4\error{0.3}    & /  &  /  &  9.8\error{1.1}  & \textbf{11.5\error{1.0}}  & /  & \textbf{9.4\error{0.2}}  & 8.1\error{0.6}   & \multirow{3}{*}{56.4\error{0.4}} \\
                                 & 10  & 15.2\error{0.5}    &  /   & /   & 20.9\error{0.8} & \textbf{21.9\error{0.8}} & /  & 28.2\error{0.8} & \textbf{32.0\error{0.5}} &                           \\
                                 & 50  & 29.7\error{0.2}  &  /   & / & 42.7\error{0.7} & \textbf{44.2\error{0.6}} & / & 42.8\error{0.2} & \textbf{45.6\error{0.5}} &                           \\  
    \end{tabular}
    \caption{Results of depth-5 ConvNet on ImageNet-1K subsets. \includegraphics[height=2ex]{images/chart_down.png} indicates worse performance than DATM. EDF achieves SOTAs on all settings compared with DD methods. Compared with SRe2L and RDED, we achieve SOTAs on 16 out of 21 settings.}
    \label{table1}
\end{table*}

\paragraph{Baselines.}
We compare two baselines: dataset distillation (DD) methods and methods utilizing knowledge distillation (Eval. w/ Knowledge Distillation)~\citep{Hinton2015DistillingTK}. 
For DD methods, we compare trajectory-matching-based methods such as MTT~\citep{Cazenavette2022DatasetDB}, FTD~\citep{Du2022MinimizingTA}, and DATM~\citep{guo2024lossless}.
In the knowledge distillation group, we compare against SRe2L~\citep{Yin2023SqueezeRA} and RDED~\citep{Sun2023OnTD}. 
The results for subsets not covered in these papers are obtained through replication using the official open-source recipe.

\subsection{Main Results}

\begin{table}[h]
    \centering
    \tablestyle{9pt}{1.1}
    
    \begin{tabular}{c|cc|cc}
    Subset & \multicolumn{2}{c|}{ImageMeow} & \multicolumn{2}{c}{ImageYellow} \\
    IPC    & 200            & 300           & 200            & 300            \\ \shline
    Random & 52.8\error{0.4}       & 55.3\error{0.3}      & 70.5\error{0.5}       & 72.8\error{0.8}       \\
    DATM    & 60.7\error{0.6}   & 64.1\error{0.7} & 79.3\error{1.0} & 82.1\error{1.1} \\
    EDF    & 62.5\error{0.7}       & \textbf{65.9\error{0.6}}      & 81.0\error{0.9}       & \textbf{84.2\error{0.7}}       \\ \shline
    Full   & \multicolumn{2}{c|}{65.2\error{1.3}}  & \multicolumn{2}{c}{83.2\error{0.9}}     \\
    \end{tabular}
    \caption{Bolded entries are lossless performances under IPC300.}
    \label{lossless}
\end{table}

\begin{table}[]
    \centering
    \tablestyle{5pt}{1.1}
    
    \begin{tabular}{cc|cc|cc}
    \multirow{2}{*}{Dataset}       & \multirow{2}{*}{IPC} & \multicolumn{2}{c|}{DD} & \multicolumn{2}{c}{Eval. w/ KD} \\
                                   &                      & DATM       & EDF        & RDED           & EDF            \\ \hline
    \multirow{3}{*}{CIFAR-10}      & 1                    & 46.9\error{0.5}   & \textbf{47.8\error{0.3}}   & 23.5\error{0.3}       & \textbf{28.1\error{0.5}}       \\
                                   & 50                   & 76.1\error{0.3}   & \textbf{77.3\error{0.5}}   & 68.4\error{0.1}       & \textbf{69.5\error{0.3}}       \\
                                   & 500                  & 83.5\error{0.2}   & \textbf{84.8\error{0.5}}   & /              & 86.1\error{0.4}       \\ \hline
    \multirow{3}{*}{Tiny-ImageNet} & 1                    & 17.1\error{0.3}   & \textbf{17.8\error{0.2}}   & \textbf{12.0\error{0.1}}       & 10.3\error{0.4}       \\
                                   & 10                   & 31.1\error{0.3}   & \textbf{32.5\error{0.5}}   & 39.6\error{0.1}       & \textbf{40.5\error{0.7}}       \\
                                   & 50                   & 39.7\error{0.3}   & \textbf{41.1\error{0.6}}   & 47.6\error{0.2}       & \textbf{48.9\error{0.3}}     
    \end{tabular}
    \caption{EDF also outperforms baselines of two approaches on datasets in simple scenarios, such as CIFAR-10 and Tiny-ImageNet.}
    \label{tab:simple_dataset}
\end{table}

\paragraph{ImageNet-1K Subsets.}
We compare our approach with previous works on various ImageNet-1K subsets.
As shown in Table~\ref{table1}, EDF consistently achieves state-of-the-art (SOTA) results across all settings. 
On larger IPCs, \textit{i.e.}, 200 or 300, the performance of EDF significantly outperforms that observed with smaller IPCs.
We achieve lossless performances on ImageMeow and ImageYellow under IPC300, 23\% of real data, as shown in Table~\ref{lossless}.
When evaluated against Eval. w/ Knowledge Distillation methods, our distilled datasets outperform SRe2L and RDED in most settings. 
It is important to note that applying knowledge distillation (KD) for evaluation tends to reduce EDF's pure dataset distillation performance, particularly in low IPC (images per class) settings such as IPC1 and IPC10. This occurs because smaller IPCs lack the capacity to effectively incorporate the knowledge from a well-trained teacher model. We also provide results without knowledge distillation in Appendix~\ref{app_wo_kd}. 

\paragraph{CIFAR-10 and Tiny-ImageNet.}
We also evaluate our approach on simple datasets, \textit{i.e.}, CIFAR-10 and Tiny-ImageNet.
As illustrated in Table~\ref{tab:simple_dataset}, EDF can still outperform the SOTAs of the DD and Eval. w/ KD approaches in most settings.
The results confirm the good generalization of our approach in simple scenarios. 

\begin{table*}[h]
\centering
\tablestyle{4pt}{1.1}
\begin{subtable}[h]{\textwidth}
        \begin{tabular}{c|cc|cc|cc|cc|cc|cc}
        Method & \multicolumn{2}{c|}{Bird-Easy} & \multicolumn{2}{c|}{Bird-Hard} & \multicolumn{2}{c|}{Dog-Easy} & \multicolumn{2}{c|}{Dog-Hard} & \multicolumn{2}{c|}{Car-Easy} & \multicolumn{2}{c}{Car-Hard} \\
        IPC    & 10               & 50              & 10               & 50              & 10              & 50              & 10              & 50              & 10              & 50              & 10              & 50             \\ \shline
        Random &      32.4\error{0.5}            & 53.8\error{0.6}       &    22.6\error{0.7}              & 41.8\error{0.5}        &  26.0\error{0.4}               & 30.8\error{0.8}        &    14.5\error{0.2}             & 27.6\error{0.7}        &    18.2\error{0.4}             &    34.4\error{0.3}             &    25.6\error{0.5}             &     40.4\error{0.5}           \\
        FTD    & 60.0\error{1.1}         & 63.4\error{0.6}        & 54.4\error{0.8}         & 59.6\error{1.2}        & 41.1\error{1.3}        & 45.9\error{0.9}        & 36.5\error{1.1}        & 43.5\error{0.9}        & 44.4\error{1.1}        & 49.6\error{0.5}        & 52.1\error{0.5}        & 55.6\error{0.9}       \\
        DATM   & 62.2\error{0.4}         & 67.1\error{0.3}        & 56.0\error{0.5}         & 62.9\error{0.8}        & 42.8\error{0.7}        & 48.2\error{0.5}        & 38.6\error{0.7}        & 47.4\error{0.5}        & 46.4\error{0.5}        & 53.8\error{0.6}        & 53.2\error{0.6}        & 58.7\error{0.8}       \\
        EDF    & \textbf{63.4\error{0.5}}         & \textbf{69.0\error{0.8}}        & \textbf{57.1\error{0.4}}         & \textbf{64.8\error{0.6}}        & \textbf{43.2\error{0.5}}        & \textbf{49.4\error{0.8}} & \textbf{39.6\error{0.9}} & \textbf{49.2\error{0.3}}  & \textbf{47.6\error{0.7}} & \textbf{54.6\error{0.2}}  & \textbf{55.4\error{0.8}} & \textbf{61.0\error{0.5}}  \\ \hline
        Full   & \multicolumn{2}{c|}{81.6\error{1.0}}      & \multicolumn{2}{c|}{82.4\error{0.8}}      & \multicolumn{2}{c|}{57.3\error{0.3}}     & \multicolumn{2}{c|}{58.4\error{0.5}}     & \multicolumn{2}{c|}{63.5\error{0.2}}     & \multicolumn{2}{c}{72.8\error{1.1}}     \\
        \end{tabular}
    \caption{EDF achieves SOTAs on Bird, Dog, and Car categories of the Comp-DD benchmark}
    
    \label{benchmark_results}
    \end{subtable}
    \vfill
    \begin{subtable}[h]{\textwidth}
        \centering
        \tablestyle{9pt}{1.1}
        \begin{tabular}{c|cccc|cccc|cccc}
            Category   & \multicolumn{4}{c|}{Bird}                                      & \multicolumn{4}{c|}{Dog}                                       & \multicolumn{4}{c}{Car}                                       \\
            IPC        & \multicolumn{2}{c}{10}               & \multicolumn{2}{c|}{50} & \multicolumn{2}{c}{10}               & \multicolumn{2}{c|}{50} & \multicolumn{2}{c}{10}               & \multicolumn{2}{c}{50} \\ \shline
            Complexity & Easy   & \multicolumn{1}{c|}{Hard}   & Easy       & Hard       & Easy   & \multicolumn{1}{c|}{Hard}   & Easy       & Hard       & Easy   & \multicolumn{1}{c|}{Hard}   & Easy       & Hard      \\ 
            Recovery ratio (\%)         & \textbf{77.7} & \multicolumn{1}{c|}{69.3} & \textbf{84.6}     & 76.9     & \textbf{75.4} & \multicolumn{1}{c|}{67.8} & \textbf{86.2}     & 84.2     & \textbf{76.2} & \multicolumn{1}{c|}{75.6} & \textbf{87.4}     & 83.7    \\ 
        \end{tabular}
    \caption{Recovery ratios of easy subsets are \textbf{higher} than that of hard subsets, aligning with the complexity metrics.}
    \label{}
    \end{subtable}
    \caption{\textbf{(a)} Partial results on Bird, Dog, and Car categories of the Complex DD Benchmark under IPC 10 and 50. \textbf{(b)} Recovery ratios of EDF on the partial Complex DD Benchmark.}
    \label{table2}
\end{table*}

\paragraph{Comp-DD Benchmark.} The results for EDF on the Bird, Car, and Dog categories from the Comp-DD Benchmark are shown in Table~\ref{table2}. EDF demonstrates superior test accuracy. As expected, the recovery ratios for easy subsets are consistently higher than those for hard subsets, confirming that the hard subsets present a greater challenge for DD methods. These results validate our complexity metrics, which effectively distinguish the varying levels of difficulty between easy and hard subsets.

\subsection{Ablation Study}
\label{sec:ablation}
In this section, we perform ablation studies to analyze key components of our approach.
Unless otherwise specified, the following results are all based on ConvNetD5.

\paragraph{Effect of Modules.}
We validate the contribution of \textbf{Discriminative Area Enhancement} (DAE) and \textbf{Common Pattern Dropout} (CPD), respectively.
As illustrated in Table~\ref{ab_mod},
both DAE and CPD significantly improve the baseline performance. DAE's biased updates toward high-activation areas 
effectively enhance the discriminative features in synthetic images. CPD, on the other hand, mitigates the negative influence of common patterns by filtering out low-loss supervision, ensuring that the synthetic images retain their discriminative properties.

\begin{table}[h]
    \centering
    \tablestyle{8pt}{1.1}
        \begin{tabular}{cc|ccc}
            DAE & CPD & ImageWoof & ImageMeow &ImageYellow \\ \shline
                       &  & 39.2  &48.9 &65.1              \\ 
             & \checkmark           &   40.3  &49.5 &66.2             \\
            \checkmark           &  & 41.1   &51.2 &67.5              \\ 
            \checkmark           & \checkmark           & \textbf{41.8}     &  \textbf{52.6} &  \textbf{68.2}       \\ 
        \end{tabular}
    \caption{Ablation results of two modules, DAE and CPD, on three ImageNet-1K subsets. Both modules bring improvements to the performance, underscoring individual efficacy.}
    \label{ab_mod}
\end{table}

\paragraph{Supervision Dropout Ratio.} The dropout ratio in CPD is critical for balancing the removal of common patterns and dataset capacity (IPC). As shown in Table~\ref{ab_dropout_ratio}, smaller IPCs benefit most from moderate dropout ratios (12.5-25\%), which filter low-loss signals while preserving important information. For larger IPCs, higher dropout ratios (37.5-50\%) improve performance, as these datasets can tolerate more aggressive filtering. However, an excessively high ratio (e.g., 75\%) reduces performance across all IPCs by discarding too much information, weakening the ability to learn.

\paragraph{Frequency of Activation Map Update.} 
\label{sec:ab_freq}
To accurately capture the evolving discriminative features in synthetic images, EDF dynamically updates the Grad-CAM activation maps at a predefined frequency. The choice of update frequency should be adjusted based on the IPC to achieve optimal performance. As shown in Table~\ref{ab_update_freq}, larger IPCs benefit from a lower update frequency, as the pixel learning rate is set lower for more stable distillation. In contrast, smaller IPCs require a higher update frequency to effectively adapt to the faster changes in the synthetic images during training.

This trend is influenced by the pixel learning rate: larger IPCs can use lower rates to ensure smooth convergence, making frequent updates unnecessary. Smaller IPCs, with limited data capacity, require higher learning rates and more frequent updates to quickly adapt to changes in discriminative areas. Thus, selecting the appropriate update frequency is essential for balancing stability and adaptability in the distillation process, depending on dataset size and complexity.

\paragraph{Strategies for Discriminative Area Enhancement.}
The Discriminative Area Enhancement (DAE) component involves two key factors: the enhancement factor $\beta$ and the threshold for activation maps. Ablation studies (Table~\ref{ab_enc_fac}) show that the best performance is achieved when $\beta$ is between 1 and 2. When $\beta < 1$, some discriminative areas are diminished rather than enhanced, as their gradient weights become $< 1$. Conversely, excessively large $\beta$ values ($\geq 10$) lead to overemphasis on certain areas, distorting the overall learning process (see Appendix~\ref{app_distort} for examples of this distortion). Therefore, $\beta$ should be reasonably controlled to balance the emphasis on discriminative regions.

Dynamic thresholding using mean activation values outperforms fixed thresholds (Table~\ref{ab_act_thd}).
This is because the mean adapts to the evolving activation maps during training, whereas a fixed threshold risks either emphasizing low-activation areas if set too low or omitting key discriminative features if set too high.

\begin{table*}[t]
    \centering
    \tablestyle{6pt}{1.0}
    \begin{subtable}{0.58\textwidth}
        \centering
        \begin{tabular}{cc|cccccc}
        \multicolumn{2}{c|}{Ratio (\%)}        & 0  & 12.5 & 25 & 37.5 & 50 & 75 \\ \shline
        \multirow{3}{*}{ImageFruit}  & 1  & \textbf{32.8} & 32.4   & 32.3 & 31.8   & 30.6 & 29.1 \\
                                     & 10 & 45.4 & 45.9   & \textbf{46.5} & 46.2   & 45.8 & 44.3 \\
                                     & 50 & 49.5 & 50.1   & 50.7 & \textbf{50.9}   & 50.6 & 49.2 \\ \hline
        \multirow{3}{*}{ImageSquawk} & 1  & \textbf{41.8} & 41.3   & 41.2 & 41.0   & 39.6 & 38.1 \\
                                     & 10 & 64.8 & 65.0   & \textbf{65.4} & 65.2   & 64.9 & 63.2 \\
                                     & 50 & 73.9 & 74.2   & 74.6 & \textbf{74.8}   & 74.5 & 72.8 \\ %
        \end{tabular}
        \caption{Within a reasonable range, the target supervision dropout ratio increases as the IPC becomes larger. Dropping too much supervision could result in losing too much information.}
        \label{ab_dropout_ratio}
    \end{subtable}
    \hfill
    \begin{subtable}{0.4\textwidth}
        \centering
        \tablestyle{6pt}{1.0}
        \begin{tabular}{cccccc}
            \multicolumn{2}{c|}{Frequency (iter.)}                          & 1    & 50   & 100  & 200  \\ \shline
            \multirow{3}{*}{ImageNette}  & \multicolumn{1}{c|}{1}  & 49.4 & \textbf{51.2} & 50.5 & 49.5 \\
                                         & \multicolumn{1}{c|}{10} & 68.4 & 69.8 & \textbf{71.0} & 70.6 \\
                                         & \multicolumn{1}{c|}{50} & 72.5 & 75.6 & 76.5 & \textbf{77.8} \\ \hline
            \multirow{3}{*}{ImageYellow} & \multicolumn{1}{c|}{1}  & 47.8 & \textbf{49.4} & 49.2 & 48.2 \\
                                         & \multicolumn{1}{c|}{10} & 66.4 & 67.8 & \textbf{68.2} & 67.2 \\
                                         & \multicolumn{1}{c|}{50} & 70.4 & 72.2 & 73.1 & \textbf{73.6} \\
            \end{tabular}
        \caption{Within a reasonable range, a higher frequency performs better on small IPCs, while larger IPCs prefer a lower frequency. }
        \label{ab_update_freq}
    \end{subtable}
    \caption{\textbf{(a)} Results of different supervision dropout ratios across various IPCs. \textbf{(b)} Results of different activation map update frequencies across various IPCs.}
    \label{fig1}
    \vspace{-1em}
\end{table*}

\begin{table*}[t]
    \centering
    \begin{subtable}{0.65\textwidth}
        \begin{subtable}{0.48\textwidth}
        \centering
        \tablestyle{4pt}{1.1}
        \begin{tabular}{c|ccccc}
        \multirow{2}{*}{IPC} & \multicolumn{5}{c}{Enhancement Factor ($\beta$)} \\
                             & 0.5    &  1       & 2       & 5       & 10       \\ \shline
        1                    & 33.4 & \textbf{34.5}    & 34.3    & 33.2    & 31.8     \\
        10                   & 50.1 & \textbf{52.6}    & 52.1    & 49.4    & 49.0     \\
        50                   & 57.8 & \textbf{59.5}    & 59.2    & 58.1    & 57.6     \\ %
        \end{tabular}
        \end{subtable}
        \hfill
        \begin{subtable}{0.48\textwidth}
            \centering
            \tablestyle{4pt}{1.1}
            \renewcommand{\arraystretch}{1.2}
            \begin{tabular}{c|ccccc}
            \multirow{2}{*}{IPC} & \multicolumn{5}{c}{Enhancement Factor ($\beta$)} \\
                                 & 0.5 & 1       & 2       & 5       & 10       \\ \shline
            1                    & 29.1 & \textbf{30.8}    & 30.5    & 30.2    & 28.8     \\
            10                   & 40.9 & 41.2    & \textbf{41.8}    & 41.0    & 40.4     \\
            50                   & 47.5 & 48.2    & \textbf{48.4}    & 48.1    & 47.2     \\ 
            \end{tabular}
        \end{subtable}
        \caption{Results on ImageMeow (left) and ImageWoof (right). ImageWoof has a higher complexity. Enhancement factor should be set within a reasonable range ($\geq 1$ and $\leq5$ in general).
        }
        \label{ab_enc_fac}
    \end{subtable}
    \hfill
    \begin{subtable}{0.32\textwidth}
        \centering
        \tablestyle{4pt}{1.1}
        \begin{tabular}{c|cccc}
        \multirow{2}{*}{IPC} & \multicolumn{4}{c}{Activation Threshold} \\
                             &  0.2      & 0.5     & 0.8   & mean   \\ \shline
        1                    &  34.2     & 34.0    & 33.8  & \textbf{34.5}           \\
        10                   &  51.2     & 52.3    & 51.5  & \textbf{52.6}         \\
        50                   &  58.0     & 59.0    & 58.4  & \textbf{59.5}         \\ 
        \end{tabular}
        \caption{Using ``mean" as a dynamic threshold gives the best performance on three IPCs.}
        \label{ab_act_thd}
    \end{subtable}
    \caption{\textbf{(a)} Ablation of the enhancement factor on ImageMeow and ImageWoof. \textbf{(b)} Ablation of the activation threshold on ImageMeow.}
    \label{fig1}
    \vspace{-1em}
\end{table*}

\section{Analysis and Discussion}

\begin{figure*}[t]
    \begin{subfigure}{0.29\textwidth}
        \centering
        \includegraphics[width=\textwidth]{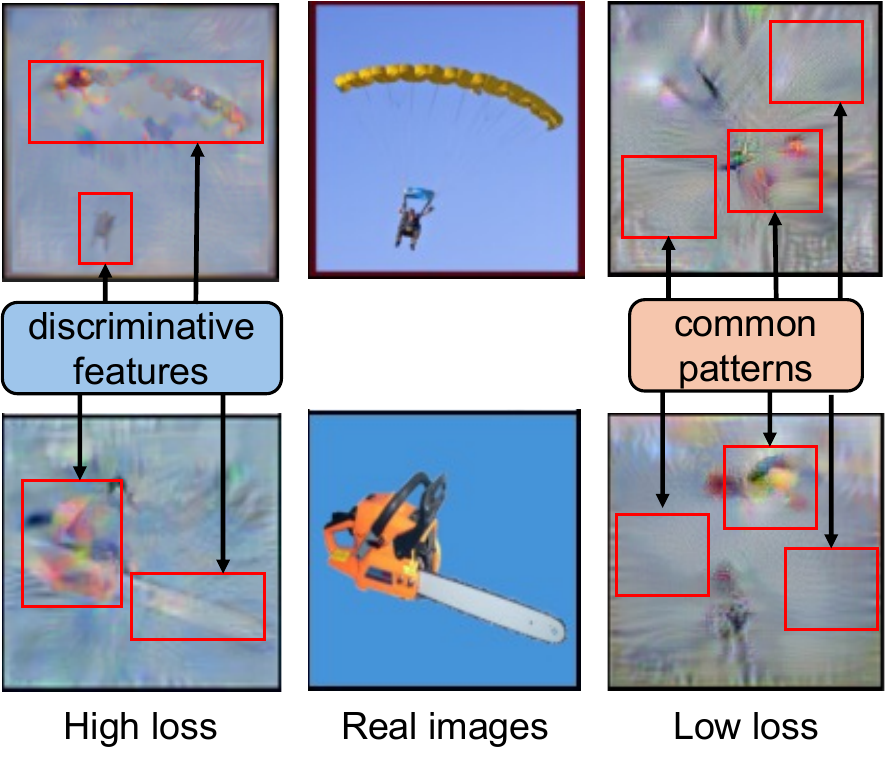}
        \caption{Low-loss supervision mainly embeds common patterns (background, colors).}
        \label{fig5a}
    \end{subfigure}
    \hfill
    \begin{subfigure}{0.69\textwidth}
        \centering
        \includegraphics[width=\textwidth]{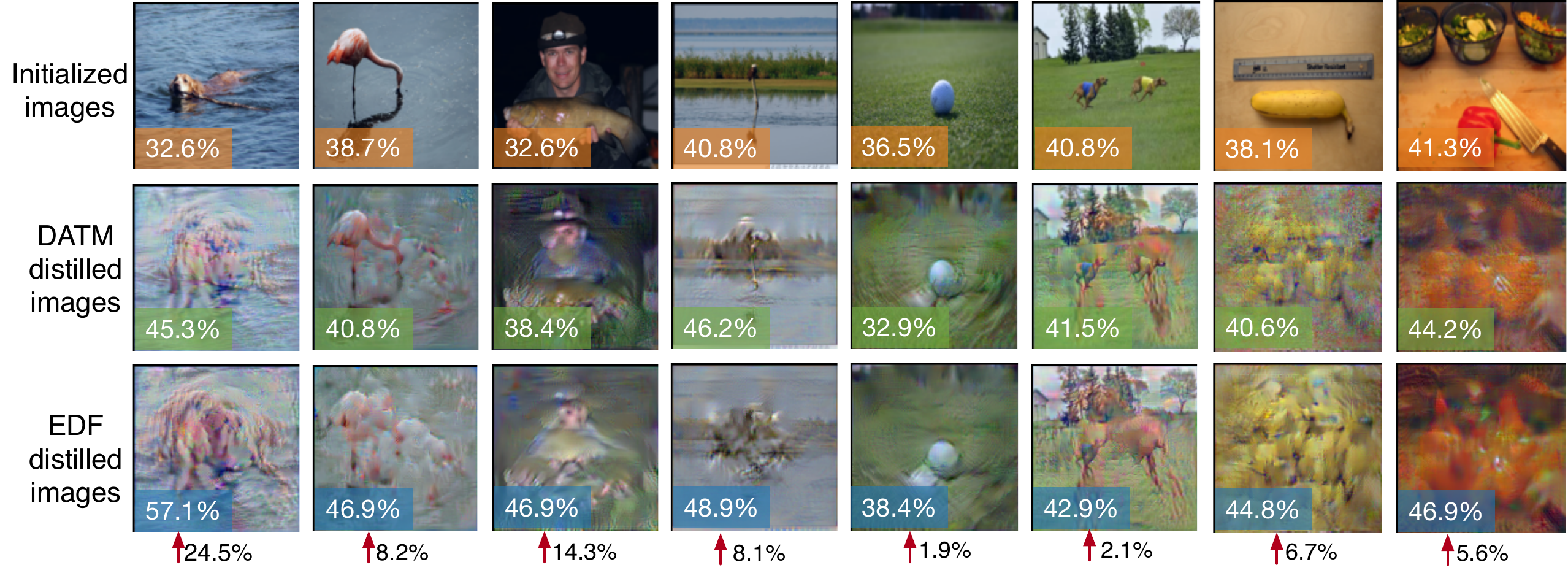}
        \caption{EDF largely increases the percentage of discriminative areas (bottom left figure of each image) with an average of 9\%, achieving the highest. Our distilled images contain more discriminative features.}
        \label{fig5b}
    \end{subfigure}
    \caption{\textbf{(a)} Comparison between high-loss and low-loss supervision distilled images. \textbf{(b)} Comparison of \textbf{discriminative areas} in images produced by initialization, DATM, and EDF. Figures at the bottom are increments made by EDF over the initial image. }
\end{figure*}

\paragraph{Disitlled Images of Different Supervision.}
As pointed out earlier, low-loss supervision tends to introduce common patterns, such as backgrounds and general colors, while high-loss supervision contains discriminative, class-specific features. 
To visualize this effect, we select two images with similar backgrounds and colors, but distinct objects.
Two images are then distilled by high-loss and low-loss supervision, respectively.
As shown in Figure~\ref{fig5a}, common patterns are indeed widely present in low-loss supervision distilled images, making two images hard to distinguish.
In contrast, high-loss supervision preserves more discriminative details, enabling the model to distinguish between two classes.
This confirms the validity of dropping low-loss supervision and underscores the effectiveness of the \textit{Common Pattern Dropout} (CPD) in mitigating the negative impact of common features.

\begin{table}[h]
    \centering
    \tablestyle{1pt}{1.1}
    \includegraphics[width=1.0\linewidth, height=0.25\linewidth]{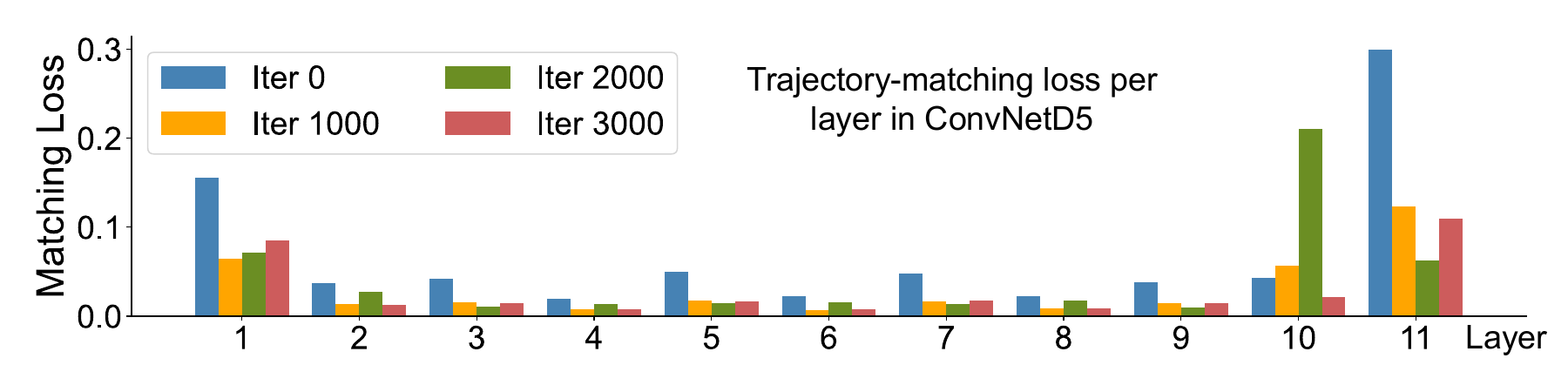}
    \begin{tabular}{c|cccccc}
        Strategy      & Random        & \multicolumn{1}{l}{Uniform} & First & Middle & Last   & EDF        \\ \shline
        Param. to layer        & 1, 2, 5, 8, 9 & 1 - 11                      & 1, 2, 3  & 5,6    & 9, 10, 11 & 2, 4, 5, 6 \\
        Acc. (\%) & 71.3          & 68.7                        & 71.9  & 73.1   & 65.4   & \textbf{74.8}       \\ 
    \end{tabular}
    \caption{EDF's loss-wise dropout performs the best. The dropping ratio of all criteria is fixed at 25\%. ``Param. to layer" refers to layers that contain dropped trajectory parameters.}
    \label{tab6}
    \vspace{-1em}
\end{table}

\paragraph{Enhancement of Discriminative Areas.}
Our Discriminative Area Enhancement (DAE) module aims to amplify updates in high-activation areas of synthetic images, as identified by Grad-CAM.
To show how DAE enhances discriminative areas, we visualize the same group of images under initialization, DATM distillation, and EDF distillation in Figure~\ref{fig5b}
We also report discriminative area statistics, computed by the percentage of pixels whose activation values are higher than the mean, on each image at the bottom left.
As can be discovered, DATM is capable of increasing discriminative regions, while EDF can achieve a more significant enhancement. 
Visually, the enhancement manifests through an increased number of core objects and enlarged areas of class-specific features.
Moreover, EDF's enhancement is more pronounced, especially when the image has smaller discriminative areas initially, e.g., discriminative features of the first column image increase by 24.5\%.
These phenomena demonstrate that EDF is effective in capturing and emphasizing discriminative features.

\paragraph{Supervision Dropout Criteria.}
We investigate the impact of the following dropout strategies: (i) dynamic dropout, which includes random selection from all layers, and (ii) static dropout, which includes uniform selection across layers and fixed selection from the first, middle, or last layers. As shown in Table~\ref{tab6}, all strategies except EDF's loss-based dropout lead to performance degradation, with uniform selection and last-layer dropout causing the most significant degradation.

The reasons for this are twofold. First, supervision of low losses—primarily located in the shallow layers of the model—is the main source of common patterns. Discarding supervision from deep layers, where loss values are higher (random selection, uniform selection, or last-layer dropout), reduces the presence of discriminative features. Second, static dropout fails to account for the dynamic nature of low-loss supervision, as trajectory-matching losses vary across layers as the distillation process evolves.

\section{Related Work}

\paragraph{Approaches.}  
DD aims to create compact datasets that maintain performance levels comparable to full-scale datasets. 
Approaches in DD can be categorized into two primary approaches: matching-based and knowledge-distillation-based.

1)~\textit{Matching-based approaches} are foundational in DD research, focusing on aligning synthetic data with real datasets by capturing essential patterns. Landmark works like gradient matching (DC)~\citep{DC}, distribution matching (DM)~\citep{DM}, and trajectory matching (MTT)~\citep{Cazenavette2022DatasetDB} extract critical metrics from real datasets, then replicate these metrics in synthetic data. Subsequent research has refined these methods, improving the fidelity of distilled datasets~\citep{DSA, Wang2022CAFELT, Zhao2023ImprovedDM, Lee2022DatasetCW, Liu2023DatasetDV, Liu2023DREAMED,  Cazenavette2023GeneralizingDD, Sajedi_2023_ICCV, khaki2024atom}. Data selection techniques have been integrated to synthesize more representative samples~\citep{Xu2023DistillGF, Sundar2023PruneTD, lee2024selmatcheffectivelyscalingdataset}. Recent advancements optimize distillation for different image-per-class (IPC) settings, balancing dataset size and information retention~\citep{Du2023SequentialSM, Chen2023DataDC, guo2024lossless, li2024prioritizealignmentdatasetdistillation, lee2024selmatcheffectivelyscalingdataset}.
Moreover, soft labels have been widely applied to improve the performance~\citep{9533769, TESLA, qin2024labelworththousandimages, yu2024heavylabelsoutdataset}.
Current matching-based methods treat pixels uniformly, overlooking discriminative features and supervision signals, limiting their effectiveness on complex datasets.

2)~\textit{Knowledge-distillation-based approaches} take an alternative route by aligning teacher-student model outputs when evaluating distilled datasets. 
Notable examples include SRe2L~\citep{Yin2023SqueezeRA} and RDED~\citep{Sun2023OnTD}.
In our work, we adopt knowledge distillation as a validation strategy for fair comparisons.

\paragraph{Benchmarks.}
Some benchmarks for dataset distillation, such as DC-BENCH~\citep{cui2022dcbenchdatasetcondensationbenchmark} and DD-Ranking~\cite{DD-Ranking}, have been proposed in light of the unfairness issue in previous DD evaluation and provide a more fair comparison scheme.
In this work, we propose the Comp-DD benchmark for DD in complex scenarios and offer a new angle to evaluate existing methods. Comp-DD systematically explores dataset distillation complexity by curating subsets from ImageNet-1K with varying degrees of difficulty. This benchmark provides a more rigorous evaluation framework, facilitating deeper exploration of DD in complex, real-world settings and encouraging further advances in the field.

\section{Conclusion}
We introduced Emphasize Discriminative Features (EDF), a dataset distillation method that enhances class-specific regions in synthetic images.
EDF addresses two key limitations of prior methods: i) enhancing discriminative regions in synthetic images using Grad-CAM activation maps, and ii) filtering out low-loss signals that embed common patterns through \textit{Common Pattern Dropout (CPD)} and \textit{Discriminative Area Enhancement (DAE)}.
EDF achieves state-of-the-art results across ImageNet-1K subsets, including lossless performance on several of them.
We also proposed the Comp-DD benchmark, designed to evaluate dataset distillation in both simple and complex settings.

\section{Acknowledgement}
We thank Bo Zhao, Guang Li, Dai Liu, Zhenghao Zhao, Songhua Liu, Ruonan Yu, Zhiyuan Liang, and Gongfan Fang for valuable discussions and feedback. This work also was supported by the National Research Foundation, Singapore, under its AI Singapore Programme (AISG Award No: AISG2-PhD-2021-08-008). Yang You’s group is being sponsored by NUS startup grant (Presidential Young Professorship), Singapore MOE Tier-1 grant, ByteDance grant, ARCTIC grant, SMI grant (WBS number: A8001104-00-00), Alibaba grant, and Google grant for TPU usage.
 {
    \small
    \bibliographystyle{ieeenat_fullname}
    \bibliography{main}

\begin{thebibliography}{78}
\providecommand{\natexlab}[1]{#1}
\providecommand{\url}[1]{\texttt{#1}}
\expandafter\ifx\csname urlstyle\endcsname\relax
  \providecommand{\doi}[1]{doi: #1}\else
  \providecommand{\doi}{doi: \begingroup \urlstyle{rm}\Url}\fi

\bibitem[Cazenavette et~al.(2022)Cazenavette, Wang, Torralba, Efros, and Zhu]{Cazenavette2022DatasetDB}
George Cazenavette, Tongzhou Wang, Antonio Torralba, Alexei~A. Efros, and Jun-Yan Zhu.
\newblock Dataset distillation by matching training trajectories.
\newblock In \emph{CVPR}, pages 10708--10717, 2022.

\bibitem[Cazenavette et~al.(2023)Cazenavette, Wang, Torralba, Efros, and Zhu]{Cazenavette2023GeneralizingDD}
George Cazenavette, Tongzhou Wang, Antonio Torralba, Alexei~A. Efros, and Jun-Yan Zhu.
\newblock Generalizing dataset distillation via deep generative prior.
\newblock In \emph{CVPR}, 2023.

\bibitem[Chen et~al.(2023{\natexlab{a}})Chen, Yang, Wang, and Mirzasoleiman]{Chen2023DataDC}
Xuxi Chen, Yu Yang, Zhangyang Wang, and Baharan Mirzasoleiman.
\newblock Data distillation can be like vodka: Distilling more times for better quality.
\newblock In \emph{ICLR}, 2023{\natexlab{a}}.

\bibitem[Chen et~al.(2024)Chen, Huang, and Weng]{chen2024provable}
Yilan Chen, Wei Huang, and Tsui-Wei Weng.
\newblock Provable and efficient dataset distillation for kernel ridge regression.
\newblock In \emph{NeurIPS}, 2024.

\bibitem[Chen et~al.(2023{\natexlab{b}})Chen, Geng, Zhu, Woisetschlaeger, Li, Schimmler, Mayer, and Rong]{chen2023comprehensivestudydatasetdistillation}
Zongxiong Chen, Jiahui Geng, Derui Zhu, Herbert Woisetschlaeger, Qing Li, Sonja Schimmler, Ruben Mayer, and Chunming Rong.
\newblock A comprehensive study on dataset distillation: Performance, privacy, robustness and fairness.
\newblock \emph{ArXiv}, 2023{\natexlab{b}}.

\bibitem[Cui et~al.(2022{\natexlab{a}})Cui, Wang, Si, and Hsieh]{TESLA}
Justin Cui, Ruochen Wang, Si Si, and Cho-Jui Hsieh.
\newblock Scaling up dataset distillation to imagenet-1k with constant memory.
\newblock In \emph{ICML}, 2022{\natexlab{a}}.

\bibitem[Cui et~al.(2022{\natexlab{b}})Cui, Wang, Si, and Hsieh]{cui2022dcbenchdatasetcondensationbenchmark}
Justin Cui, Ruochen Wang, Si Si, and Cho-Jui Hsieh.
\newblock Dc-bench: Dataset condensation benchmark.
\newblock In \emph{NeurIPS}, 2022{\natexlab{b}}.

\bibitem[Du et~al.(2022)Du, Jiang, Tan, Zhou, and Li]{Du2022MinimizingTA}
Jiawei Du, Yiding Jiang, Vincent Y.~F. Tan, Joey~Tianyi Zhou, and Haizhou Li.
\newblock Minimizing the accumulated trajectory error to improve dataset distillation.
\newblock In \emph{CVPR}, 2022.

\bibitem[Du et~al.(2023)Du, Shi, and Zhou]{Du2023SequentialSM}
Jiawei Du, Qin Shi, and Joey~Tianyi Zhou.
\newblock Sequential subset matching for dataset distillation.
\newblock \emph{ArXiv}, abs/2311.01570, 2023.

\bibitem[Du et~al.(2024)Du, Zhang, Hu, Huang, and Zhou]{du2024diversitydrivensynthesisenhancingdataset}
Jiawei Du, Xin Zhang, Juncheng Hu, Wenxin Huang, and Joey~Tianyi Zhou.
\newblock Diversity-driven synthesis: Enhancing dataset distillation through directed weight adjustment.
\newblock In \emph{NeurIPS}, 2024.

\bibitem[Fernandez(2020)]{torchcam}
François-Guillaume Fernandez.
\newblock Torchcam: class activation explorer.
\newblock \url{https://github.com/frgfm/torch-cam}, 2020.

\bibitem[Gu et~al.(2023)Gu, Vahidian, Kungurtsev, Wang, Jiang, You, and Chen]{Gu2023EfficientDD}
Jianyang Gu, Saeed Vahidian, Vyacheslav Kungurtsev, Haonan Wang, Wei Jiang, Yang You, and Yiran Chen.
\newblock Efficient dataset distillation via minimax diffusion.
\newblock \emph{ArXiv}, abs/2311.15529, 2023.

\bibitem[Guo et~al.(2024)Guo, Wang, Cazenavette, Li, Zhang, and You]{guo2024lossless}
Ziyao Guo, Kai Wang, George Cazenavette, Hui Li, Kaipeng Zhang, and Yang You.
\newblock Towards lossless dataset distillation via difficulty-aligned trajectory matching.
\newblock In \emph{ICLR}, 2024.

\bibitem[He et~al.(2015)He, Zhang, Ren, and Sun]{He2015DeepRL}
Kaiming He, X. Zhang, Shaoqing Ren, and Jian Sun.
\newblock Deep residual learning for image recognition.
\newblock In \emph{CVPR}, 2015.

\bibitem[Hinton et~al.(2015)Hinton, Vinyals, and Dean]{Hinton2015DistillingTK}
Geoffrey~E. Hinton, Oriol Vinyals, and Jeffrey Dean.
\newblock Distilling the knowledge in a neural network.
\newblock \emph{ArXiv}, abs/1503.02531, 2015.

\bibitem[Howard(2019)]{howard2019imagenette}
Jeremy Howard.
\newblock Imagenette and imagewoof: Imagenet subsets for classification.
\newblock \url{https://github.com/fastai/imagenette}, 2019.

\bibitem[Izzo and Zou(2023)]{izzo2023a}
Zachary Izzo and James Zou.
\newblock A theoretical study of dataset distillation.
\newblock In \emph{NeurIPS}, 2023.

\bibitem[Khaki et~al.(2024)Khaki, Sajedi, Wang, Liu, Lawryshyn, and Plataniotis]{khaki2024atom}
Samir Khaki, Ahmad Sajedi, Kai Wang, Lucy~Z. Liu, Yuri~A. Lawryshyn, and Konstantinos~N. Plataniotis.
\newblock Atom: Attention mixer for efficient dataset distillation.
\newblock \emph{ArXiv}, 2024.

\bibitem[Krizhevsky(2009)]{Krizhevsky2009LearningML}
Alex Krizhevsky.
\newblock Learning multiple layers of features from tiny images.
\newblock 2009.

\bibitem[Krizhevsky et~al.(2012)Krizhevsky, Sutskever, and Hinton]{Krizhevsky2012ImageNetCW}
Alex Krizhevsky, Ilya Sutskever, and Geoffrey~E. Hinton.
\newblock Imagenet classification with deep convolutional neural networks.
\newblock \emph{Communications of the ACM}, 60:\penalty0 84 -- 90, 2012.

\bibitem[Le and Yang(2015)]{Le2015TinyIV}
Ya Le and Xuan~S. Yang.
\newblock Tiny imagenet visual recognition challenge.
\newblock 2015.

\bibitem[Lee et~al.(2022{\natexlab{a}})Lee, Chun, Jung, Yun, and Yoon]{DCC}
Saehyung Lee, Sanghyuk Chun, Sangwon Jung, Sangdoo Yun, and Sung-Hoon Yoon.
\newblock Dataset condensation with contrastive signals.
\newblock In \emph{ICML}, 2022{\natexlab{a}}.

\bibitem[Lee et~al.(2022{\natexlab{b}})Lee, Chun, Jung, Yun, and Yoon]{Lee2022DatasetCW}
Saehyung Lee, Sanghyuk Chun, Sangwon Jung, Sangdoo Yun, and Sung-Hoon Yoon.
\newblock Dataset condensation with contrastive signals.
\newblock In \emph{ICML}, 2022{\natexlab{b}}.

\bibitem[Lee and Chung(2024)]{lee2024selmatcheffectivelyscalingdataset}
Yongmin Lee and Hye~Won Chung.
\newblock Selmatch: Effectively scaling up dataset distillation via selection-based initialization and partial updates by trajectory matching.
\newblock In \emph{ICML}, 2024.

\bibitem[Li et~al.(2024{\natexlab{a}})Li, Guo, Zhao, Zhang, Cheng, Khaki, Zhang, Sajedi, Plataniotis, Wang, and You]{li2024prioritizealignmentdatasetdistillation}
Zekai Li, Ziyao Guo, Wangbo Zhao, Tianle Zhang, Zhi-Qi Cheng, Samir Khaki, Kaipeng Zhang, Ahmad Sajedi, Konstantinos~N Plataniotis, Kai Wang, and Yang You.
\newblock Prioritize alignment in dataset distillation.
\newblock \emph{ArXiv}, 2024{\natexlab{a}}.

\bibitem[Li et~al.(2024{\natexlab{b}})Li, Zhong, Liang, Zhou, Shi, Wang, Zhao, Zhao, Wang, Qin, Liu, Zhang, Zhou, Zhu, Wang, Li, Zhang, Liu, Huang, Lyu, Lv, Jin, Akata, Gu, Vedantam, Shou, Deng, Yan, Shang, Cazenavette, Wu, Cui, Chen, Yao, Kellis, Plataniotis, Zhao, Wang, You, and Wang]{DD-Ranking}
Zekai Li, Xinhao Zhong, Zhiyuan Liang, Yuhao Zhou, Mingjia Shi, Ziqiao Wang, Wangbo Zhao, Xuanlei Zhao, Haonan Wang, Ziheng Qin, Dai Liu, Kaipeng Zhang, Tianyi Zhou, Zheng Zhu, Kun Wang, Guang Li, Junhao Zhang, Jiawei Liu, Yiran Huang, Lingjuan Lyu, Jiancheng Lv, Yaochu Jin, Zeynep Akata, Jindong Gu, Rama Vedantam, Mike Shou, Zhiwei Deng, Yan Yan, Yuzhang Shang, George Cazenavette, Xindi Wu, Justin Cui, Tianlong Chen, Angela Yao, Manolis Kellis, Konstantinos~N. Plataniotis, Bo Zhao, Zhangyang Wang, Yang You, and Kai Wang.
\newblock Dd-ranking: Rethinking the evaluation of dataset distillation.
\newblock GitHub repository, 2024{\natexlab{b}}.

\bibitem[Liu et~al.(2024)Liu, Gu, Cao, Trinitis, and Schulz]{ATT}
Dai Liu, Jindong Gu, Hu Cao, Carsten Trinitis, and Martin Schulz.
\newblock Dataset distillation by automatic training trajectories.
\newblock In \emph{ECCV}, 2024.

\bibitem[Liu et~al.(2023{\natexlab{a}})Liu, Xing, Li, Dalal, He, and Wang]{Liu2023DatasetDV}
Haoyang Liu, Tiancheng Xing, Luwei Li, Vibhu Dalal, Jingrui He, and Haohan Wang.
\newblock Dataset distillation via the wasserstein metric.
\newblock \emph{ArXiv}, abs/2311.18531, 2023{\natexlab{a}}.

\bibitem[Liu et~al.(2022)Liu, Wang, Yang, Ye, and Wang]{liu2022dataset}
Songhua Liu, Kai Wang, Xingyi Yang, Jingwen Ye, and Xinchao Wang.
\newblock Dataset distillation via factorization.
\newblock In \emph{NeurIPS}, 2022.

\bibitem[Liu et~al.(2023{\natexlab{b}})Liu, Gu, Wang, Zhu, Jiang, and You]{Liu2023DREAMED}
Yanqing Liu, Jianyang Gu, Kai Wang, Zheng~Hua Zhu, Wei Jiang, and Yang You.
\newblock Dream: Efficient dataset distillation by representative matching.
\newblock \emph{ICCV}, pages 17268--17278, 2023{\natexlab{b}}.

\bibitem[Loo et~al.(2022)Loo, Hasani, Amini, and Rus]{RFAD}
Noel Loo, Ramin Hasani, Alexander Amini, and Daniela Rus.
\newblock Efficient dataset distillation using random feature approximation.
\newblock In \emph{NeurIPS}, 2022.

\bibitem[Loo et~al.(2023)Loo, Hasani, Lechner, and Rus]{RCIG}
Noel Loo, Ramin Hasani, Mathias Lechner, and Daniela Rus.
\newblock Dataset distillation with convexified implicit gradients.
\newblock In \emph{ICML}, 2023.

\bibitem[Loo et~al.(2024)Loo, Maalouf, Hasani, Lechner, Amini, and Rus]{loo2024large}
Noel Loo, Alaa Maalouf, Ramin Hasani, Mathias Lechner, Alexander Amini, and Daniela Rus.
\newblock Large scale dataset distillation with domain shift.
\newblock In \emph{ICML}, 2024.

\bibitem[Lorraine et~al.(2019)Lorraine, Vicol, and Duvenaud]{lorraine2019optimizingmillionshyperparametersimplicit}
Jonathan Lorraine, Paul Vicol, and David Duvenaud.
\newblock Optimizing millions of hyperparameters by implicit differentiation.
\newblock In \emph{PMLR}, 2019.

\bibitem[Maalouf et~al.(2023)Maalouf, Tukan, Loo, Hasani, Lechner, and Rus]{maalouf2023sizeapproximationerrordistilled}
Alaa Maalouf, Murad Tukan, Noel Loo, Ramin Hasani, Mathias Lechner, and Daniela Rus.
\newblock On the size and approximation error of distilled sets.
\newblock In \emph{NeurIPS}, 2023.

\bibitem[Miao et~al.(2024)Miao, Liu, Zhao, Guo, Yang, Zheng, and Jensen]{miao2024moreefficienttimeseries}
Hao Miao, Ziqiao Liu, Yan Zhao, Chenjuan Guo, Bin Yang, Kai Zheng, and Christian~S. Jensen.
\newblock Less is more: Efficient time series dataset condensation via two-fold modal matching--extended version.
\newblock \emph{arXiv}, 2024.

\bibitem[Moser et~al.(2024)Moser, Raue, Palacio, Frolov, and Dengel]{Moser2024LatentDD}
Brian~B. Moser, Federico Raue, Sebasti{\'a}n~M. Palacio, Stanislav Frolov, and Andreas Dengel.
\newblock Latent dataset distillation with diffusion models.
\newblock \emph{ArXiv}, abs/2403.03881, 2024.

\bibitem[Nguyen et~al.(2021{\natexlab{a}})Nguyen, Chen, and Lee]{KIP-FC}
Timothy Nguyen, Zhourong Chen, and Jaehoon Lee.
\newblock Dataset meta-learning from kernel ridge-regression.
\newblock In \emph{ICLR}, 2021{\natexlab{a}}.

\bibitem[Nguyen et~al.(2021{\natexlab{b}})Nguyen, Novak, Xiao, and Lee]{KIP-ConvNet}
Timothy Nguyen, Roman Novak, Lechao Xiao, and Jaehoon Lee.
\newblock Dataset distillation with infinitely wide convolutional networks.
\newblock In \emph{NeurIPS}, 2021{\natexlab{b}}.

\bibitem[Qin et~al.(2024)Qin, Deng, and Alvarez-Melis]{qin2024labelworththousandimages}
Tian Qin, Zhiwei Deng, and David Alvarez-Melis.
\newblock A label is worth a thousand images in dataset distillation.
\newblock In \emph{NeurIPS}, 2024.

\bibitem[Sajedi et~al.(2023)Sajedi, Khaki, Amjadian, Liu, Lawryshyn, and Plataniotis]{Sajedi_2023_ICCV}
Ahmad Sajedi, Samir Khaki, Ehsan Amjadian, Lucy~Z. Liu, Yuri~A. Lawryshyn, and Konstantinos~N. Plataniotis.
\newblock Datadam: Efficient dataset distillation with attention matching.
\newblock In \emph{ICCV}, pages 17097--17107, 2023.

\bibitem[Selvaraju et~al.(2016)Selvaraju, Das, Vedantam, Cogswell, Parikh, and Batra]{Selvaraju2016GradCAMVE}
Ramprasaath~R. Selvaraju, Abhishek Das, Ramakrishna Vedantam, Michael Cogswell, Devi Parikh, and Dhruv Batra.
\newblock Grad-cam: Visual explanations from deep networks via gradient-based localization.
\newblock \emph{International Journal of Computer Vision}, 128:\penalty0 336 -- 359, 2016.

\bibitem[Shang et~al.(2023)Shang, Yuan, and Yan]{shang2023mim4ddmutualinformationmaximization}
Yuzhang Shang, Zhihang Yuan, and Yan Yan.
\newblock Mim4dd: Mutual information maximization for dataset distillation.
\newblock In \emph{NeurIPS}, 2023.

\bibitem[Shin et~al.(2023)Shin, Bae, Shin, Joo, and Moon]{LCMat}
Seung-Jae Shin, Heesun Bae, DongHyeok Shin, Weonyoung Joo, and Il-Chul Moon.
\newblock Loss-curvature matching for dataset selection and condensation.
\newblock In \emph{AISTATS}, 2023.

\bibitem[Simonyan and Zisserman(2014)]{Simonyan2014VeryDC}
Karen Simonyan and Andrew Zisserman.
\newblock Very deep convolutional networks for large-scale image recognition.
\newblock \emph{CoRR}, abs/1409.1556, 2014.

\bibitem[Su et~al.(2024)Su, Hou, Gao, Tian, and Tang]{D4M}
Duo Su, Junjie Hou, Weizhi Gao, Yingjie Tian, and Bowen Tang.
\newblock D4m: Dataset distillation via disentangled diffusion model.
\newblock In \emph{CVPR}, 2024.

\bibitem[Sucholutsky and Schonlau(2021)]{9533769}
Ilia Sucholutsky and Matthias Schonlau.
\newblock Soft-label dataset distillation and text dataset distillation.
\newblock In \emph{2021 International Joint Conference on Neural Networks (IJCNN)}, pages 1--8, 2021.

\bibitem[Sun et~al.(2024)Sun, Shi, Yu, and Lin]{Sun2023OnTD}
Peng Sun, Bei Shi, Daiwei Yu, and Tao Lin.
\newblock On the diversity and realism of distilled dataset: An efficient dataset distillation paradigm.
\newblock In \emph{CVPR}, 2024.

\bibitem[Sundar et~al.(2023)Sundar, Keskin, Chandak, Chen, Ghahremani, and Ghosh]{Sundar2023PruneTD}
Anirudh~S. Sundar, G{\"o}kçe Keskin, Chander Chandak, I-Fan Chen, Pegah Ghahremani, and Shalini Ghosh.
\newblock Prune then distill: Dataset distillation with importance sampling.
\newblock In \emph{ICASSP}, 2023.

\bibitem[Vicol et~al.(2022)Vicol, Lorraine, Pedregosa, Duvenaud, and Grosse]{pmlr-v162-vicol22a}
Paul Vicol, Jonathan~P Lorraine, Fabian Pedregosa, David Duvenaud, and Roger~B Grosse.
\newblock On implicit bias in overparameterized bilevel optimization.
\newblock In \emph{PMLR}, 2022.

\bibitem[Wang et~al.(2022)Wang, Zhao, Peng, Zhu, Yang, Wang, Huang, Bilen, Wang, and You]{Wang2022CAFELT}
Kai Wang, Bo Zhao, Xiangyu Peng, Zheng~Hua Zhu, Shuo Yang, Shuo Wang, Guan Huang, Hakan Bilen, Xinchao Wang, and Yang You.
\newblock Cafe learning to condense dataset by aligning features.
\newblock In \emph{CVPR}, 2022.

\bibitem[Wang et~al.(2023)Wang, Gu, Zhou, Zhu, Jiang, and You]{Wang2023DiMDD}
Kai Wang, Jianyang Gu, Daquan Zhou, Zheng~Hua Zhu, Wei Jiang, and Yang You.
\newblock Dim: Distilling dataset into generative model.
\newblock \emph{ArXiv}, abs/2303.04707, 2023.

\bibitem[Wang et~al.(2025)Wang, Yang, Liu, Sun, Hu, He, and Zhang]{wang2025datasetdistillationneuralcharacteristic}
Shaobo Wang, Yicun Yang, Zhiyuan Liu, Chenghao Sun, Xuming Hu, Conghui He, and Linfeng Zhang.
\newblock Dataset distillation with neural characteristic function: A minmax perspective.
\newblock In \emph{CVPR}, 2025.

\bibitem[Wang et~al.(2020)Wang, Zhu, Torralba, and Efros]{wang2020dataset}
Tongzhou Wang, Jun-Yan Zhu, Antonio Torralba, and Alexei~A. Efros.
\newblock Dataset distillation.
\newblock \emph{ArXiv}, 2020.

\bibitem[Wei et~al.(2023)Wei, Cao, Yang, and Ma]{wei2023sparse}
Xing Wei, Anjia Cao, Funing Yang, and Zhiheng Ma.
\newblock Sparse parameterization for epitomic dataset distillation.
\newblock In \emph{NeurIPS}, 2023.

\bibitem[Wu et~al.(2024{\natexlab{a}})Wu, Zhang, Deng, and Russakovsky]{wu2024visionlanguagedatasetdistillation}
Xindi Wu, Byron Zhang, Zhiwei Deng, and Olga Russakovsky.
\newblock Vision-language dataset distillation.
\newblock In \emph{TMLR}, 2024{\natexlab{a}}.

\bibitem[Wu et~al.(2024{\natexlab{b}})Wu, Du, Liu, Lin, Xu, and Cheng]{wu2024ddrobustbenchadversarialrobustnessbenchmark}
Yifan Wu, Jiawei Du, Ping Liu, Yuewei Lin, Wei Xu, and Wenqing Cheng.
\newblock Dd-robustbench: An adversarial robustness benchmark for dataset distillation.
\newblock \emph{ArXiv}, 2024{\natexlab{b}}.

\bibitem[Xiao and He(2024)]{xiao2024largescalesoftlabelsnecessary}
Lingao Xiao and Yang He.
\newblock Are large-scale soft labels necessary for large-scale dataset distillation?
\newblock In \emph{NeurIPS}, 2024.

\bibitem[Xu et~al.(2023)Xu, Li, Cui, Wang, Lu, Tai, and Tang]{Xu2023DistillGF}
Yue Xu, Yong-Lu Li, Kaitong Cui, Ziyu Wang, Cewu Lu, Yu-Wing Tai, and Chi-Keung Tang.
\newblock Distill gold from massive ores: Efficient dataset distillation via critical samples selection.
\newblock \emph{ArXiv}, abs/2305.18381, 2023.

\bibitem[Xu et~al.(2024)Xu, Lin, Qiu, Lu, and Li]{xu2024lowranksimilarityminingmultimodal}
Yue Xu, Zhilin Lin, Yusong Qiu, Cewu Lu, and Yong-Lu Li.
\newblock Low-rank similarity mining for multimodal dataset distillation.
\newblock In \emph{ICML}, 2024.

\bibitem[Yang et~al.(2024{\natexlab{a}})Yang, Cheng, Hong, Fan, Wei, and Liu]{NSD}
Shaolei Yang, Shen Cheng, Mingbo Hong, Haoqiang Fan, Xing Wei, and Shuaicheng Liu.
\newblock Neural spectral decomposition for dataset distillation.
\newblock In \emph{ECCV}, 2024{\natexlab{a}}.

\bibitem[Yang et~al.(2024{\natexlab{b}})Yang, Zhu, Deng, and Russakovsky]{yang2024datasetdistillationlearning}
William Yang, Ye Zhu, Zhiwei Deng, and Olga Russakovsky.
\newblock What is dataset distillation learning?
\newblock In \emph{ICML}, 2024{\natexlab{b}}.

\bibitem[Yin and Shen(2023)]{yin2023datasetdistillationlargedata}
Zeyuan Yin and Zhiqiang Shen.
\newblock Dataset distillation in large data era.
\newblock \emph{ArXiv}, 2023.

\bibitem[Yin et~al.(2023)Yin, Xing, and Shen]{Yin2023SqueezeRA}
Zeyuan Yin, Eric~P. Xing, and Zhiqiang Shen.
\newblock Squeeze, recover and relabel: Dataset condensation at imagenet scale from a new perspective.
\newblock In \emph{NeurIPS}, 2023.

\bibitem[Yu et~al.(2024)Yu, Liu, Chen, Ye, and Wang]{yu2024heavylabelsoutdataset}
Ruonan Yu, Songhua Liu, Zigeng Chen, Jingwen Ye, and Xinchao Wang.
\newblock Heavy labels out! dataset distillation with label space lightening.
\newblock \emph{ArXiv}, 2024.

\bibitem[Yuan et~al.(2024)Yuan, Wang, Baktashmotlagh, Luo, and Huang]{yuan2024colororiented}
Bowen Yuan, Zijian Wang, Mahsa Baktashmotlagh, Yadan Luo, and Zi Huang.
\newblock Color-oriented redundancy reduction in dataset distillation.
\newblock In \emph{NeurIPS}, 2024.

\bibitem[Zhang et~al.(2023)Zhang, Li, Wang, Zeng, and Ge]{M3D}
Hansong Zhang, Shikun Li, Pengju Wang, Dan Zeng, and Shiming Ge.
\newblock M3d: Dataset condensation by minimizing maximum mean discrepancy.
\newblock In \emph{AAAI}, 2023.

\bibitem[Zhao and Bilen(2021{\natexlab{a}})]{DM}
Bo Zhao and Hakan Bilen.
\newblock Dataset condensation with distribution matching.
\newblock \emph{WACV}, pages 6503--6512, 2021{\natexlab{a}}.

\bibitem[Zhao and Bilen(2021{\natexlab{b}})]{DSA}
Bo Zhao and Hakan Bilen.
\newblock Dataset condensation with differentiable siamese augmentation.
\newblock In \emph{ICML}, 2021{\natexlab{b}}.

\bibitem[Zhao and Bilen(2022)]{zhao2022synthesizinginformativetrainingsamples}
Bo Zhao and Hakan Bilen.
\newblock Synthesizing informative training samples with gan, 2022.

\bibitem[Zhao et~al.(2021)Zhao, Mopuri, and Bilen]{DC}
Bo Zhao, Konda~Reddy Mopuri, and Hakan Bilen.
\newblock Dataset condensation with gradient matching.
\newblock In \emph{ICML}, 2021.

\bibitem[Zhao et~al.(2023)Zhao, Li, Qin, and Yu]{Zhao2023ImprovedDM}
Ganlong Zhao, Guanbin Li, Yipeng Qin, and Yizhou Yu.
\newblock Improved distribution matching for dataset condensation.
\newblock \emph{CVPR}, pages 7856--7865, 2023.

\bibitem[Zhao et~al.(2024{\natexlab{a}})Zhao, Han, Tang, Li, Song, Wang, Wang, and You]{zhao2024stitch}
Wangbo Zhao, Yizeng Han, Jiasheng Tang, Zhikai Li, Yibing Song, Kai Wang, Zhangyang Wang, and Yang You.
\newblock A stitch in time saves nine: Small vlm is a precise guidance for accelerating large vlms.
\newblock \emph{ArXiv}, 2024{\natexlab{a}}.

\bibitem[Zhao et~al.(2024{\natexlab{b}})Zhao, Tang, Han, Song, Wang, Huang, Wang, and You]{zhao2024dynamic}
Wangbo Zhao, Jiasheng Tang, Yizeng Han, Yibing Song, Kai Wang, Gao Huang, Fan Wang, and Yang You.
\newblock Dynamic tuning towards parameter and inference efficiency for vit adaptation.
\newblock In \emph{NeurIPS}, 2024{\natexlab{b}}.

\bibitem[Zhao et~al.(2024{\natexlab{c}})Zhao, Shang, Wu, and Yan]{zhao2024datasetquantizationactivelearning}
Zhenghao Zhao, Yuzhang Shang, Junyi Wu, and Yan Yan.
\newblock Dataset quantization with active learning based adaptive sampling.
\newblock In \emph{ECCV}, 2024{\natexlab{c}}.

\bibitem[Zhao et~al.(2025)Zhao, Wang, Shang, Wang, and Yan]{zhao2025distillinglongtaileddatasets}
Zhenghao Zhao, Haoxuan Wang, Yuzhang Shang, Kai Wang, and Yan Yan.
\newblock Distilling long-tailed datasets.
\newblock In \emph{CVPR}, 2025.

\bibitem[Zhong et~al.(2024)Zhong, Fang, Chen, Gu, Dai, Qiu, and Xia]{H-GLaD}
Xinhao Zhong, Hao Fang, Bin Chen, Xulin Gu, Tao Dai, Meikang Qiu, and Shu-Tao Xia.
\newblock Hierarchical features matter: A deep exploration of gan priors for improved dataset distillation.
\newblock \emph{ArXiv}, abs/2406.05704, 2024.

\bibitem[Zhou et~al.(2022)Zhou, Nezhadarya, and Ba]{FRePo}
Yongchao Zhou, Ehsan Nezhadarya, and Jimmy Ba.
\newblock Dataset distillation using neural feature regression.
\newblock In \emph{NeurIPS}, 2022.

\end{thebibliography}
}

\clearpage
\setcounter{page}{1}
\maketitlesupplementary
We organize our supplementary material as follows:

\paragraph{Algorithm of EDF:}
\begin{itemize}
    \item Appendix~\ref{sec:pseudo_code}: Pseudo code of EDF with detailed explanation.
\end{itemize}

\paragraph{Experimental Settings:}
\begin{itemize}
\item Appendix~\ref{app_exp_set_train}: Training recipe.
\item Appendix~\ref{app_exp_set_eval}: Evaluation recipe.
\item Appendix~\ref{app_exp_set_comp}: Computing resources required for different settings.
\end{itemize}

\paragraph{Additional Experimental Results and Findings:}
\begin{itemize}
\item Appendix~\ref{app_cross_arch}: Cross-architecture evaluation.
\item Appendix~\ref{app_wo_kd}: Results of distilled datasets without knowledge-distillation-based evaluation.
\item Appendix~\ref{app_distort}: Distorted synthetic images under excessive enhancement factors.
\item Appendix~\ref{app_trend}: The changing trend of discriminative areas in EDF distillation.
\item Appendix~\ref{app_cam_models}: The impact of using different models to extract activation maps. 
\end{itemize}

\paragraph{Comp-DD Benchmark}
\begin{itemize}
\item Appendix~\ref{app_comp_dd_subset}: Subset details of the Comp-DD benchmark.
\item Appendix~\ref{app_comp_dd_param}: Hyper-parameters of the Comp-DD benchmark.
\item Appendix~\ref{app_comp_dd_metrics}: More clarifications on the complexity metrics.
\end{itemize}

\paragraph{Visualization}
\begin{itemize}
    \item Appendix~\ref{app_visual}: Visualization of EDF distilled images.
\end{itemize}

\paragraph{Related Work}
\begin{itemize}
    \item Appendix~\ref{app_rel_work}: More related work of dataset distillation.
\end{itemize}

\section{Algorithm of EDF}
\label{sec:pseudo_code}
\begin{algorithm}[htb]
    \caption{Emphasizing Discriminative Features}
    \label{algo}
    \textbf{Input:} $D_{real}$: The real dataset \\
    \textbf{Input:} $D_{syn}$: The synthetic dataset \\
    \textbf{Input:} $\mathcal{A}$: A trajectory-matching based algorithm \\
    \textbf{Input:} $\mathcal{G}$: Grad-CAM model \\ 
    \textbf{Input:} $K$: Activation maps update frequency \\
    \textbf{Input:} $\alpha$: Threshold of supervision dropout \\
    \textbf{Input:} $T$: Total distillation steps \\
    \textbf{Input:} $\beta$: Enhancement factor \\
    \textbf{Input:} $\mathcal{F}$: Activation map processing function \\
    \textbf{Input:} $r$: Learning rate of synthetic dataset \\
    \begin{algorithmic}[1]
    
    \FOR{$t$ in $0\dots T-1$}
        
        \STATE $L \leftarrow \mathcal{A}(D_{syn}, D_{real})$ \ \ \ \ \ \ \ \ \ \ \ \ \ \ \ \ \ \ \ \ \ \ \ \ \ \ \ \ \ \ \ensuremath{\triangleright} Compute the array of trajectory matching losses

        \STATE $L' \leftarrow Sort(L)$ \ \ \ \ \ \ \ \ \ \ \ \ \ \ \ \ \ \ \ \ \ \ \ \ \ \ \ \ \ \ \ \ \  \ \ \ \ \ \ \ \  \ensuremath{\triangleright} Sort $L$ to get ordered losses

        \STATE $L_{edf} \leftarrow \sum_{i=\alpha |L|}^{|L|} L'_{i}$ \ \ \ \ \ \ \ \ \ \ \ \ \ \ \ \ \ \ \ \ \ \ \ \ \ \ \ \ \ \ \ \ \ensuremath{\triangleright} Dropout low-loss supervision

        \IF{$t \bmod K = 0$}
            \STATE $M \leftarrow \mathcal{G}(D_{syn})$ \ \ \ \ \ \ \ \ \ \ \ \ \ \ \ \ \ \ \ \ \ \ \ \ \ \ \ \ \ \ \ \ \ \ \ \ \ensuremath{\triangleright} Update activation maps of current $S$
        \ENDIF

        \STATE $(\nabla D_{syn})_{EDF} \leftarrow \nabla D_{syn} \circ \mathcal{F}(M, \beta)$ \ \ \ \ \ \ensuremath{\triangleright} Process synthetic image gradients

        \STATE $D_{syn} \leftarrow D_{syn} - r\cdot (\nabla D_{syn})_{EDF}$ \ \ \ \ \ \ \ \ \ \ensuremath{\triangleright} Biased update towards discriminative areas
    \ENDFOR
    \STATE Return $D_{syn}$
    \end{algorithmic}
\end{algorithm}

Algorithm~\ref{algo} provides a pseudo-code of EDF.
Lines 1-7 specify inputs of the EDF, including a trajectory-matching algorithm $\mathcal{A}$, the model for Grad-CAM $\mathcal{G}$, the frequency of activation map update $K$, the supervision dropout ratio $\alpha$, the enhancement factor $\beta$, the activation map processing function $\mathcal{F}$, and the number of distillation iterations $T$.

Lines 12-14 describe the Common Pattern Dropout module.
After we obtain the trajectory matching losses from $\mathcal{A}$, we sort them in ascending order to get ordered losses.
Then, the smallest $\alpha |L|$ elements are dropped as they introduce non-discriminative common patterns.

Lines 15-19 describe the Discriminative Area Enhancement module.
For every $K$ iterations, we update activation maps of synthetic images.
The gradients of synthetic images are then processed by the function $\mathcal{F}$ (see Equation~\ref{eq4} for the computation).
Finally, synthetic images are updated biasedly towards discriminative areas.

\section{Experimental Settings}

\subsection{Training Details}
\label{app_exp_set_train}
We follow previous trajectory matching works~\citep{Du2022MinimizingTA, guo2024lossless, li2024prioritizealignmentdatasetdistillation} to train expert trajectories for one hundred epochs. 
Hyper-parameters are directly adopted without modification.
For distillation, we implement EDF based on DATM~\citep{guo2024lossless} and PAD~\citep{li2024prioritizealignmentdatasetdistillation}, which simultaneously distills soft labels along with images.

We use torch-cam~\citep{torchcam} for Grad-CAM implementation.
Hyper-parameters are listed in Table~\ref{table6}.

\subsection{Evaluation Details}
\label{app_exp_set_eval}
To achieve a fair comparison, when comparing EDF with DD methods, we only adopt the set of differentiable augmentations commonly used in previous studies~\citep{DSA, DM, Cazenavette2022DatasetDB} to train a surrogate model on distilled data and labels.

When comparing EDF with DD+KD methods, we follow their evaluation methods, which we detail the steps as follows:
\begin{enumerate}
    \item Train a teacher model on the real dataset and freeze it afterward.
    \item Train a student model on the distilled dataset by \textbf{minimizing the KL-Divergence loss} between the output of the student model and the output of the teacher model on the same batch from distilled data. 
    \item Validate the student model on the test set and obtain test accuracy.
\end{enumerate}
For implementation, please refer to the official repo of SRe2L\footnote{\url{https://github.com/VILA-Lab/SRe2L/tree/main/SRe2L}} and RDED\footnote{\url{https://github.com/LINs-lab/RDED}}.

\subsection{Computing Resources}
\label{app_exp_set_comp}
Experiments on IPC 1/10 can be run with 4x Nvidia-A100 80GB GPUs, and experiments on IPC 50 can be run with 8x Nvidia-A100 80GB GPUs.
The GPU memory demand is primarily dictated by the volume of synthetic data per batch and the total training iterations the augmentation model undergoes with that data. 
When IPC becomes large, GPU usage can be optimized by either adopting techniques like TESLA~\citep{TESLA} or by scaling down the number of training iterations ("syn\_steps") or shrinking the synthetic data batch size ("batch\_syn").

\begin{table*}[t]
\centering
\tablestyle{15pt}{1.2}
\begin{tabular}{cc|c|cc|cccc}
\multicolumn{2}{c|}{Modules}          & CPD      & \multicolumn{2}{c|}{DAE} & \multicolumn{4}{c}{TM}                                                                   \\
\multicolumn{2}{c|}{Hyper-parameters} & $\alpha$ & $\beta$       & $K$      & $T$                    & batch\_syn & lr\_pixel  & \multicolumn{1}{l}{syn\_steps} \\ \shline
\multirow{3}{*}{ImageNette}    & 1    & 0        & 1             & 50       & \multirow{3}{*}{10000} & 1000      & 10000     & 40           \\
                               & 10   & 0.25     & 1             & 100      &                        & 250       & 1000       &    40                          \\
                               & 50   & 0.375    & 1             & 200      &                        & 100       & 100   &   80                            \\ 
\multirow{3}{*}{ImageWoof}     & 1    & 0        & 1             & 50       & \multirow{3}{*}{10000} & 1000      & 10000    & 40           \\
                               & 10   & 0.25     & 1             & 100      &                        & 250       & 1000      &   40                            \\
                               & 50   & 0.375    & 2             & 200      &                        & 100       & 100        &    80                           \\
\multirow{3}{*}{ImageMeow}     & 1    & 0        & 1             & 50       & \multirow{3}{*}{10000} & 1000      & 10000      & 40           \\
                               & 10   & 0.25     & 1             & 100      &                        & 250       & 1000     &   40                            \\
                               & 50   & 0.375    & 2             & 200      &                        & 200       & 100       &     40                          \\
\multirow{3}{*}{ImageYellow}   & 1    & 0        & 1             & 50       & \multirow{3}{*}{10000} & 1000      & 10000       & 40           \\
                               & 10   & 0.25     & 1             & 100      &                        & 250       & 1000       &  40                             \\
                               & 50   & 0.375    & 1             & 200      &                        & 200       & 100  &  40                             \\
\multirow{3}{*}{ImageFruit}    & 1    & 0        & 1             & 50       & \multirow{3}{*}{10000} & 1000      & 10000        & 40           \\
                               & 10   & 0.25     & 1             & 100      &                        & 250       &  1000     &   40                            \\
                               & 50   & 0.375    & 1             & 200      &                        & 200       &  100     &  40                             \\
\multirow{3}{*}{ImageSquawk}   & 1    & 0        & 1             & 50       & \multirow{3}{*}{10000} & 1000      &  10000      & 40           \\
                               & 10   & 0.25     & 1             & 100      &                        & 250       &  1000     &   40                            \\
                               & 50   & 0.375    & 2             & 200      &                        & 100       &   100     &   80                            \\
\end{tabular}
\caption{Hyper-parameters of experiments on ImageNet nette, woof, meow, fruit, yellow, squawk subsets.}
\label{table6}
\end{table*}

\section{Additional experiment results and findings}
\subsection{Cross-architecture Evaluation}
\label{app_cross_arch}
Generalizability on different model architectures is one key property of a well-distilled dataset.
To show that EDF can generalize well on different models, we evaluate synthetic images under IPC 10 and 50 of the ImageSquawk subset, on three other standard models, AlexNet~\citep{Krizhevsky2012ImageNetCW}, VGG11~\citep{Simonyan2014VeryDC}, and ResNet18~\citep{He2015DeepRL}.
As shown in Table~\ref{table7}, our distilled datasets outperform random selection and two baseline methods on both IPC10 and IPC50.
Compared with IPC10, distilled images under IPC50 can achieve better performance on unseen neural networks.
This suggests that EDF's distillation results have decent generalizability across different architectures, especially when the compressing ratio is smaller which allows distilled datasets to accommodate more discriminative information.

\begin{table*}[]
    \centering
    \begin{subtable}{0.47\textwidth}
        \centering
        \tablestyle{8pt}{1.2}
        \begin{tabular}{c|cccc}
        Method & ConvNetD5 & ResNet18 & VGG11 & AlexNet \\ \shline
        Random &  41.8         &  40.9    & 43.2  &  35.7   \\
        FTD    &  62.8         &  49.8        & 50.5      &  47.6       \\
        DATM   &  65.1         &  \textbf{52.4}        & 51.2      & \textbf{49.6}        \\
        \textbf{EDF} & \textbf{68.2} &  50.8    & \textbf{53.2}  & 48.2    \\ 
        \end{tabular}
        \caption{ImageYellow, IPC10}
        \label{cross_arch_yellow}
    \end{subtable}
    \hfill
    \begin{subtable}{0.47\textwidth}
        \centering
        \tablestyle{8pt}{1.2}
        \begin{tabular}{c|cccc}
        Method & ConvNetD5   & ResNet18 & VGG11   & AlexNet \\ \shline
        Random &  29.6        & 31.4     & 30.8     & 25.7       \\
        FTD    &  58.4        & 55.6     & 57.6     & 52.3   \\
        DATM   &  61.8        & 62.8     & \textbf{65.6}     & 63.5   \\
        \textbf{EDF}   & \textbf{65.4} & \textbf{63.6} & 64.8 & \textbf{69.2} \\
        \end{tabular}
        
        \caption{ImageSquawk, IPC50}
        \label{cross_arch_squawk}
    \end{subtable}
    \caption{Cross-architecture evaluation on ResNet18, VGG11, and AlexNet. ConvNetD5 is the distillation architecture. Distilled datasets under IPC10 and IPC50 outperform random selection, FTD, and DATM, showing good generalizability.}
    \label{table7}
\end{table*}

\subsection{Eval. without Knowledge Distillation}
\label{app_wo_kd}
Starting from~\cite{wang2020dataset}, representative dataset distillation (DD) methods~\citep{DC, DSA, Cazenavette2022DatasetDB, Wang2022CAFELT} establish a general workflow as follows: 1) \textit{Distillation}: At this stage, information from the real dataset is fully accessible to the DD algorithm to train synthetic data. 2) \textit{Evaluation}: After the distilled dataset is obtained, the evaluation is performed by training a randomly initialized model on the distilled data. Specifically, in the context of classification, the objective is to minimize cross-entropy loss.
Recently, some new methods~\citep{Yin2023SqueezeRA, Sun2023OnTD, zhao2024dynamic, zhao2024stitch} introduced teacher knowledge into the student model by applying knowledge distillation.
Although it helps improve performances to a large extent, it may not be able to reflect the effectiveness of dataset distillation accurately.

To this end, we remove the knowledge distillation from Eval. w/ Knowledge Distillation (SRe2L and RDED) methods but keep soft labels to ensure a fair comparison,
Specifically, we train a classification model on the synthetic images by only minimizing the cross-entropy loss between student output and soft labels. 
As shown in Table~\ref{tab9}, without knowledge distillation, EDF outperforms SRe2L and RDED in 8 out of 9 settings.
Our advantage is more pronounced, especially when IPC is smaller, underscoring the superior efficacy of EDF on smaller compressing ratios.
\begin{table*}[t]
    \centering
    \tablestyle{10pt}{1.2}
    \begin{tabular}{cl|ccc|ccc|ccc}
    \multicolumn{2}{c|}{Dataset} & \multicolumn{3}{c|}{ImageNette} & \multicolumn{3}{c|}{ImageWoof} & \multicolumn{3}{c}{ImageSquawk} \\
    \multicolumn{2}{c|}{IPC}     & 1         & 10       & 50       & 1        & 10       & 50       & 1         & 10       & 50       \\ \shline
    \multicolumn{2}{c|}{Random}   & 12.6±1.5  & 44.8±1.3 & 60.4±1.4 & 11.4±1.3 & 20.2±1.2 & 28.2±0.9 & 13.2±1.1  & 29.6±1.5 & 52.8±0.4 \\
    \multicolumn{2}{c|}{DM}   & 28.2±1.5  & 58.1±1.1 & 65.8±1.1 & 19.6±1.4 & 30.4±1.3 & 36.3±1.4 & 19.7±1.3  & 30.0±1.0 & 61.5±1.2 \\
    \multicolumn{2}{c|}{MTT}   & 47.7±0.9  & 63.0±1.3 & 69.2±1.0 & 28.6±0.8 & 35.8±1.8 & 42.3±0.9 & 39.4±1.5  & 52.3±1.0 & 65.4±1.2 \\
    \multicolumn{2}{c|}{SRe2L}   & 18.4±0.8  & 41.0±0.3 & 55.6±0.2 & 16.0±0.2 & 32.2±0.3 & 35.8±0.2 & 22.5±0.5  & 35.6±0.4 & 42.2±0.3 \\
    \multicolumn{2}{c|}{RDED}    & 28.0±0.5  & 53.6±0.8 & 72.8±0.3 & 19.0±0.3 & 32.6±0.5 & \textbf{52.6±0.6} & 33.8±0.5  & 52.2±0.5 & 71.6±0.8 \\
    \multicolumn{2}{c|}{EDF}     & \textbf{52.6±0.5}  & \textbf{71.0±0.8} & \textbf{77.8±0.5} & \textbf{30.8±1.0} & \textbf{41.8±0.2} & 48.4±0.5 & \textbf{41.8±0.5}  & \textbf{65.4±0.8} & \textbf{74.8±1.2} \\
    \end{tabular}
    \caption{Performances of SRe2L and RDED without using knowledge distillation during evaluation. EDF outperforms the other two methods in most of settings, and our advantage is more pronounced as IPC gets smaller.}
    \label{tab9}
\end{table*}

\begin{table}[]
\tablestyle{8pt}{1.2}
\begin{tabular}{c|cccc}
\multirow{2}{*}{IPC} & \multicolumn{4}{c}{Grad-CAM Model}                \\
                     & ConvNetD5 & ResNet18      & ResNet50      & VGG11 \\ \shline
1                    & 52.3      & \textbf{52.6} & 52.5          & 52.5  \\
10                   & \textbf{71.2}      & 71.0          & 70.8 & 70.7  \\
50                   & 77.4      & \textbf{77.8} & 77.6          & 77.6 
\end{tabular}
\caption{Results of using different Grad-CAM models on ImageNette. The choice of model only has minor influence on the performance.}
\label{tab10}
\end{table}

\begin{table}[]
\tablestyle{10pt}{1.2}
\begin{tabular}{c|ccccc}
IPC         & 1    & 10   & 50   & 200  & 300 \\ \shline
Latency (s) & 0.63 & 0.52 & 0.74 & 0.68 & 0.94
\end{tabular}
\caption{Latency of extracting Grad-CAM activation maps using ResNet18. For each IPC in our experiments, the latency is less than one second.}
\label{tab11}
\end{table}

\subsection{Distorted Images of Large Enhancement Factor}
\label{app_distort}
\begin{figure}[htb]
    \centering
    \includegraphics[width=0.9\linewidth]{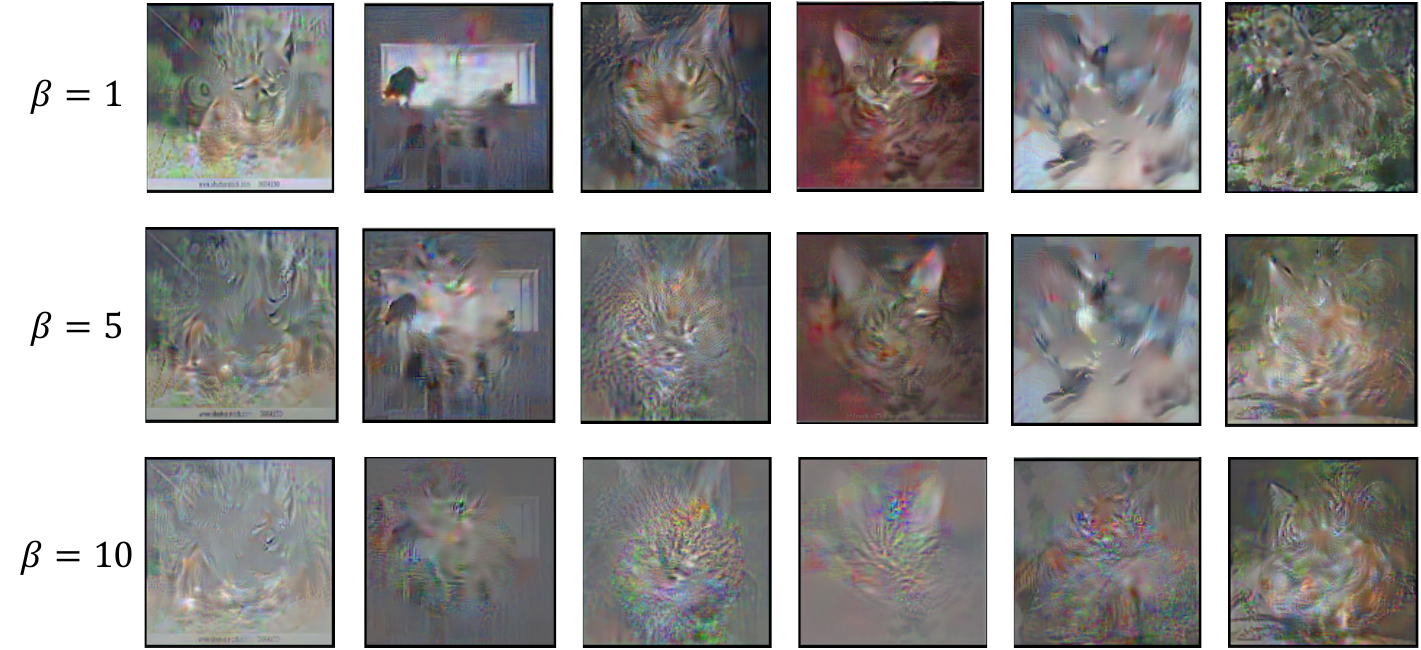}
    \caption{Distorted image distributions due to excessively large enhancement factors ($= 10$)}
    \label{fig7}
\end{figure}

In Figure~\ref{fig7}, we show results of using excessively large enhancement factors as mentioned in Section~\ref{sec:ablation}.
The distributions of these distilled images are distorted, with many pixels containing only blurred information.
This occurs because excessively increasing the gradients in discriminative areas can lead to large updates between iterations, resulting in the divergence of the pixel distribution.
Therefore, the enhancement of discrimination areas is not the stronger the better.
It is important to maintain the enhancement factor within a reasonable range.

\subsection{Changing Trends of Discriminative Areas}
\label{app_trend}
\begin{figure}[htb]
    \centering
    \begin{subfigure}{0.47\textwidth}
        \centering
        \includegraphics[width=\linewidth]{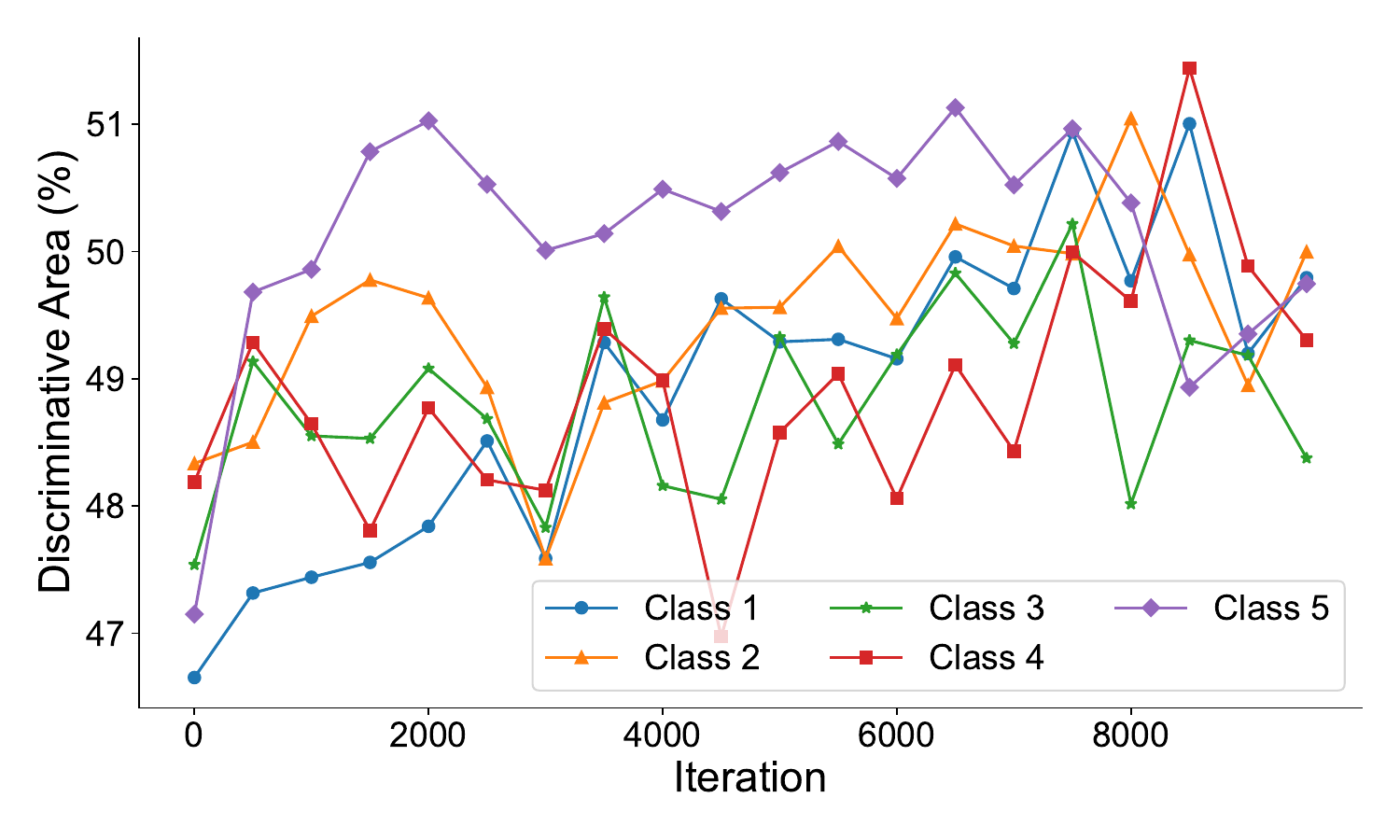}
        \caption{In general, discriminative areas show a trend of increase as the distillation proceeds.}
        \label{act_trend}
    \end{subfigure}
    \vfill
    \begin{subfigure}{0.33\textwidth}
        \centering
        \includegraphics[width=\textwidth]{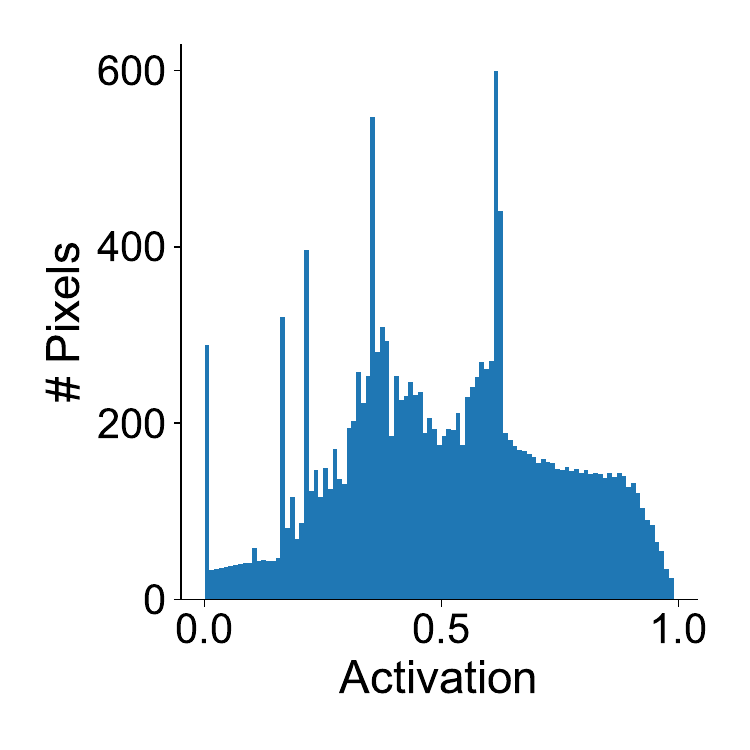}
        \caption{Most of the pixels have activation around 0.25 to 0.75.}
        \label{act_dist}
    \end{subfigure}
    \caption{\textbf{(a)} The trend of discriminative area change across various distillation iterations. \textbf{(b)} Distribution of the activation map of a random image from ImageNet-1K.}
\end{figure}

Figure~\ref{fig5b} demonstrates that EDF effectively expands the discriminative areas (high-activation regions) on several image samples at a fixed distillation iteration.
In Figure~\ref{act_trend}, we show the changing trend of discriminative areas of 5 different classes across 10000 iterations (sampled every 500 iterations).
Despite the fluctuation, these areas expand as the distillation proceeds.
This further confirms that one key factor in the success of EDF is that it successfully increases discriminative features in synthetic images and turns them into more informative data samples.

\subsection{Impact of different Grad-CAM Models}
\label{app_cam_models}
In our experiments, we use ResNe18 as the Grad-CAM model to extract activation maps.
However, the choice of Grad-CAM model does not have a significant impact on the performance, as long as it has been trained on the target dataset.
As shown in Table~\ref{tab10}, differences between performances of different Grad-CAM models are within 0.5, demonstrating that our discriminative area enhancement module doesn't depend on the choice of Grad-CAM model.

\subsection{Latency of Grad-CAM}
We use torchcam~\cite{torchcam} implementation of various Grad-CAM methods to extract activation maps.
In Table~\ref{tab11}, we show the latencies of extracting activation maps of various IPCs.
Notably, these latencies are all below one second, demonstrating a high inference speed. 
In our experiments, the maximum number of extractions for one distillation is 200 (on IPC1). 
Thus, the total time used for activation map extraction is at most two minutes, which is neglectable compared with the latency of the full distillation (several hours). 
In conclusion, our use of Grad-CAM activation maps to provide guidance doesn't reduce the efficiency of the backbone. 

\section{Comp-DD Benchmark}

\subsection{Subset Details}
\label{app_comp_dd_subset}
The corresponding class labels for each subset are listed as follows:
\begin{itemize}
    \item \textbf{Bird-Hard:} n01537544, n01592084, n01824575, n01558993, n01534433, n01843065, n01530575, n01560419, n01601694, n01532829
    \item \textbf{Bird-Easy:} n02007558, n02027492, n01798484, n02033041, n02012849, n02025239, n01818515, n01820546, n02051845, n01608432
    \item \textbf{Dog-Hard:} n02107683, n02107574, n02109525, n02096585, n02085620, n02113712, n02086910, n02093647, n02086079, n02102040
    \item \textbf{Dog-Easy:} n02096294, n02093428, n02105412, n02089973, n02109047, n02109961, n02105056, n02092002, n02114367, n02110627
    \item \textbf{Car-Hard:} n04252077, n03776460, n04335435, n03670208, n03594945, n03445924, n03444034, n04467665, n03977966, n02704792
    \item \textbf{Car-Easy:} n03459775, n03208938, n03930630, n04285008, n03100240, n02814533, n03770679, n04065272, n03777568, n04037443
    \item \textbf{Snake-Hard:} n01693334, n01687978, n01685808, n01682714, n01688243, n01737021, n01751748, n01739381, n01728920, n01728572
    \item \textbf{Snake-Easy:} n01749939, n01735189, n01729977, n01734418, n01742172, n01744401, n01756291, n01755581, n01729322, n01740131
    \item \textbf{Insect-Hard:} n02165456, n02281787, n02280649, n02172182, n02281406, n02165105, n02264363, n02268853, n01770081, n02277742
    \item \textbf{Insect-Easy:} n02279972, n02233338, n02219486, n02206856, n02174001, n02190166, n02167151, n02231487, n02168699, n02236044
    \item \textbf{Fish-Hard:} n01440764, n02536864, n02514041, n02641379, n01494475, n02643566, n01484850, n02640242, n01698640, n01873310
    \item \textbf{Fish-Easy:} n01496331, n01443537, n01498041, n02655020, n02526121, n01491361, n02606052, n02607072, n02071294, n02066245
    \item \textbf{Round-Hard:} n04409515, n04254680, n03982430, n04548280, n02799071, n03445777, n03942813, n03134739, n04039381, n09229709
    \item \textbf{Round-Easy:} n02782093, n03379051, n07753275, n04328186, n02794156, n09835506, n02802426, n04540053, n04019541, n04118538
    \item \textbf{Music-Hard:} n02787622, n03495258, n02787622, n03452741, n02676566, n04141076, n02992211, n02672831, n03272010, n03372029
    \item \textbf{Music-Easy:} n03250847, n03854065, n03017168, n03394916, n03721384, n03110669, n04487394, n03838899, n04536866, n04515003
\end{itemize}

\subsection{Complexity Metrics}
\label{app_comp_dd_metrics}
We use the percentage of pixels whose Grad-CAM activation values exceed a predefined fixed threshold to evaluate the complexity of an image.
In our settings, the fixed threshold is 0.5.
The reasons for fixing the threshold at 0.5 are twofold.
Firstly, when selecting subsets, images are static and won't be updated in any form (this is different from EDF's DAE module, which updates synthetic images).
Thus, using a fixed threshold is sufficient for determining the high-activation areas.

Secondly, values of a Grad-CAM activation map range from $0$ to $1$, with higher values corresponding to higher activation.
We present the distribution of the activation map of a random image from ImageNet-1K in Figure~\ref{}, where the majority of pixels have activation values between 0.25 and 0.75.
Subsequently, if the threshold is too small or too large, the complexity scores of all classes will be close (standard deviation is small), as shown in Figure~\ref{fig9} and~\ref{fig10}.
This results in no clear distinguishment between easy and hard subsets. 
Finally, we set 0.5 as the threshold, which is the middle point of the range.
Complexity distribution under this threshold is shown in Figure~\ref{fig11}.

Our complexity metrics are an early effort to define how complex an image is in the context of dataset distillation.
We acknowledge potential biases or disadvantages and encourage future studies to continue the refinement of complex metrics. 

\begin{figure}[t]
    \centering
    \includegraphics[width=\linewidth]{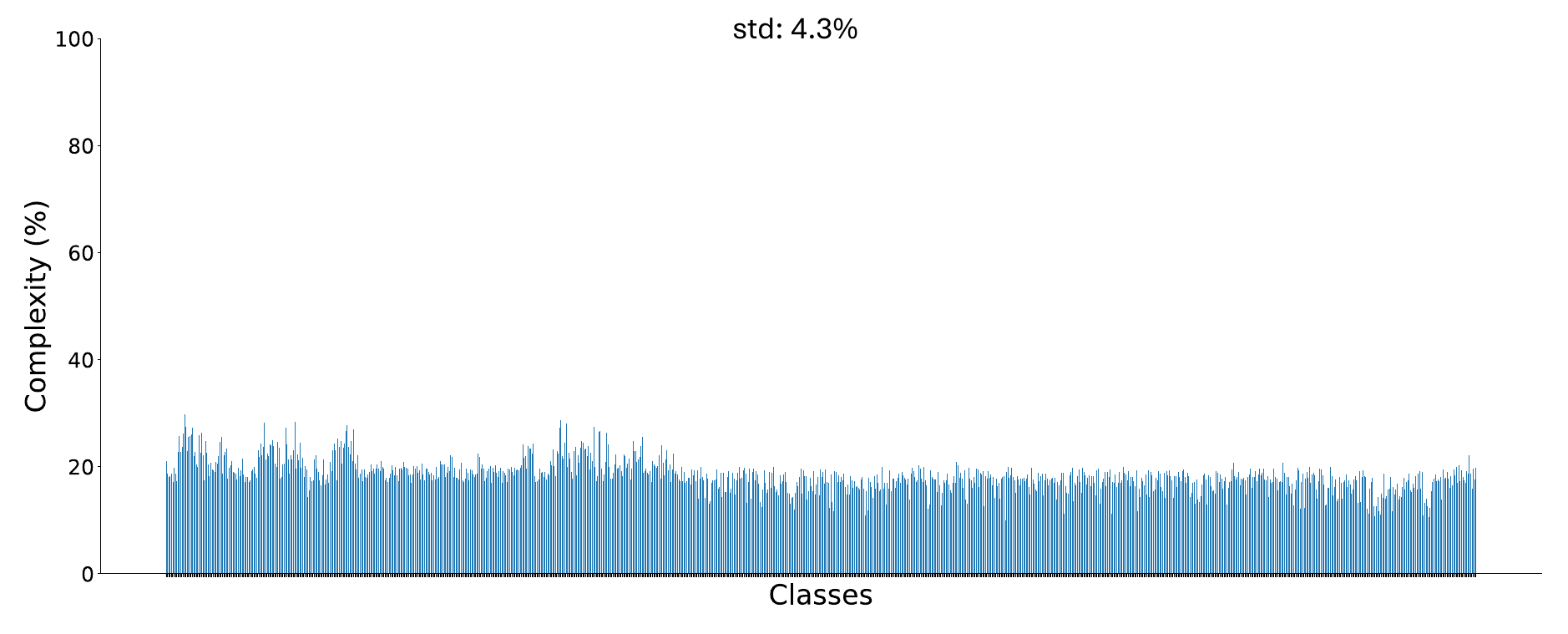}
    \caption{Complexity distribution of all classes from ImageNet-1K under threshold being 0.1. An excessively small threshold will cause the complexity of all classes to become low and difficult to distinguish.}
    \label{fig9}
\end{figure}

\begin{figure}[t]
    \centering
    \includegraphics[width=\linewidth]{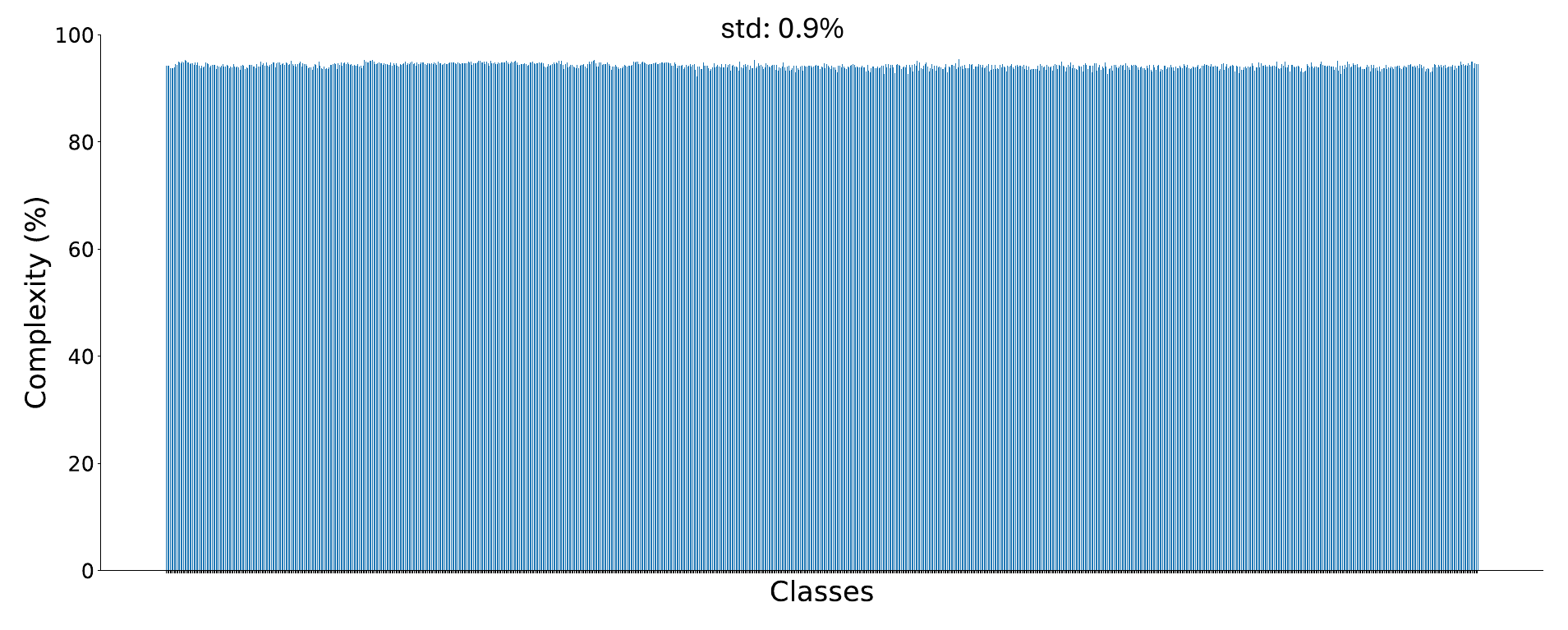}
    \caption{Complexity distribution of all classes from ImageNet-1K under threshold being 0.9. An excessively large threshold will cause the complexity of all classes to become high and difficult to distinguish.}
    \label{fig10}
\end{figure}

\begin{figure}[t]
    \centering
    \includegraphics[width=\linewidth]{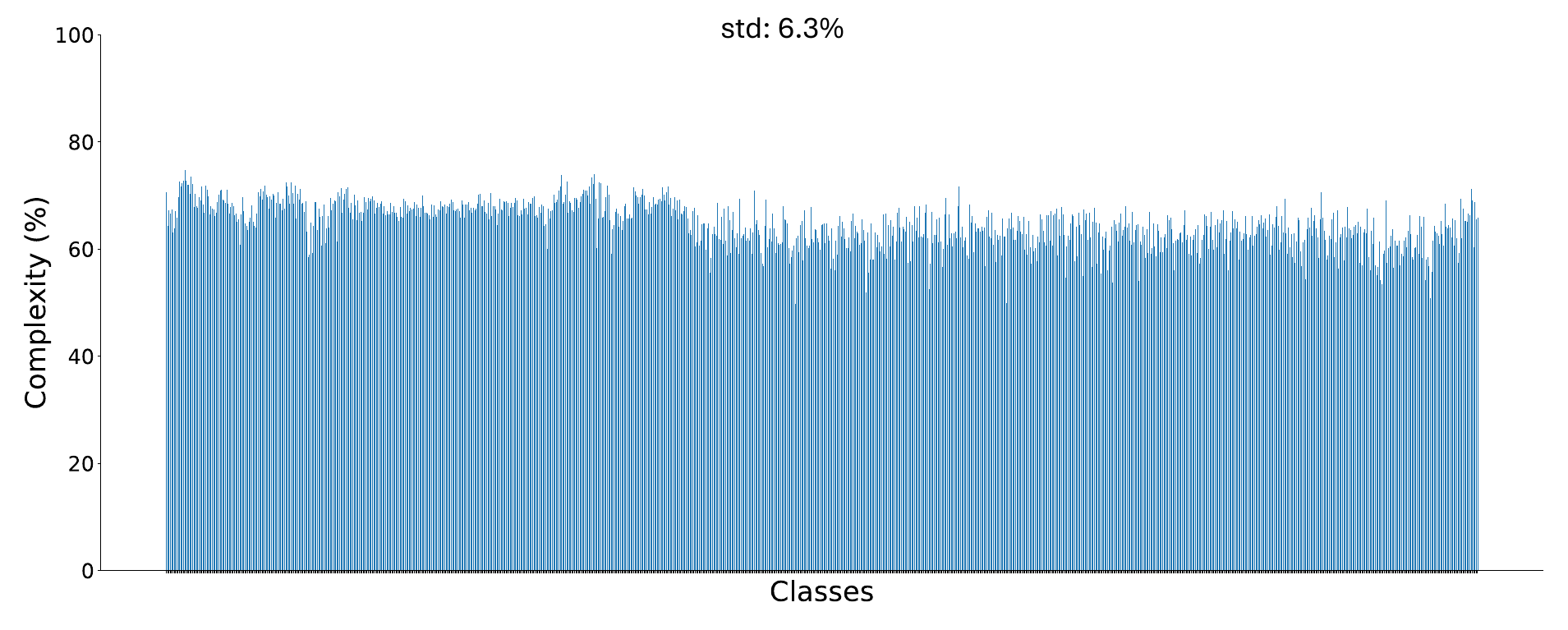}
    \caption{Complexity distribution of all classes from ImageNet-1K under threshold being 0.5. A moderate threshold makes the complexity differences between classes more distinct.}
    \label{fig11}
\end{figure}

\subsection{Benchmark Hyper-parameters}
\label{app_comp_dd_param}
For the trajectory training, experiment settings are the same as those used for ImageNet-1K and its subsets.
For distillation, we provide hyper-parameters of EDF on the Complex DD Benchmark in Table~\ref{table7}.
These hyper-parameters can serve as a reference for future works to extend to other subsets of the benchmark.

\begin{table*}[t]
\centering
\tablestyle{15pt}{1.2}
\begin{tabular}{cc|c|cc|cccc}
\multicolumn{2}{c|}{Modules}          & CPD      & \multicolumn{2}{c|}{DAE} & \multicolumn{4}{c}{TM}                                                                                       \\
\multicolumn{2}{c|}{Hyper-parameters} & $\alpha$ & $\beta$       & $K$      & $T$                    & batch\_syn & lr\_pixel & \multicolumn{1}{l}{syn\_steps} \\ \shline
\multirow{3}{*}{CDD-Bird-Easy}  & 1   & 0        & 1             & 50       & \multirow{3}{*}{10000} & 1000      & 10000     & \multirow{3}{*}{40}           \\
                                & 10  & 0.25     & 1             & 100      &                        & 400       & 1000       &                               \\
                                & 50  & 0.375    & 2             & 200      &                        & 200       & 100                          &                               \\
\multirow{3}{*}{CDD-Bird-Hard}  & 1   & 0        & 1             & 50       & \multirow{3}{*}{10000} & 1000      & 10000        & \multirow{3}{*}{40}           \\
                                & 10  & 0.25     & 1             & 100      &                        & 400       & 1000    &      \\
                                & 50  & 0.375    & 2             & 200      &                        & 200       & 100     &   \\
\multirow{3}{*}{CDD-Dog-Easy}   & 1   & 0        & 1             & 50       & \multirow{3}{*}{10000} & 1000      & 10000         & \multirow{3}{*}{40}           \\
                                & 10  & 0.25     & 1             & 100      &                        & 400       & 1000  &  \\
                                & 50  & 0.375    & 2    & 200      &       & 200  & 100   &   \\
\multirow{3}{*}{CDD-Dog-Hard}   & 1   & 0        & 1             & 50       & \multirow{3}{*}{10000} & 1000      & 10000        & \multirow{3}{*}{40}           \\
                                & 10  & 0.25     & 1             & 100      &                        & 400       & 1000    &                               \\
                                & 50  & 0.375    & 2             & 200      &                        & 200       & 100    &                               \\ 
\multirow{3}{*}{CDD-Car-Easy}   & 1   & 0        & 1             & 50       & \multirow{3}{*}{10000} & 1000      & 10000  & \multirow{3}{*}{40}           \\
                                & 10  & 0.25     & 1             & 100      &                        & 400       & 1000   &                               \\
                                & 50  & 0.375    & 2             & 200      &                        & 200       & 100   &                               \\ 
\multirow{3}{*}{CDD-Car-Hard}   & 1   & 0        & 1             & 50       & \multirow{3}{*}{10000} & 1000      & 10000  & \multirow{3}{*}{40}           \\
                                & 10  & 0.25     & 1             & 100      &                        & 400       & 1000  &                               \\
                                & 50  & 0.375    & 2             & 200      &                        & 200       & 100    &    \\ 
\end{tabular}
\caption{Hyper-parameters of EDF on the Complex DD Benchmark.}
\label{table7}
\end{table*}

\section{Visualization of Distilled Images on ImageNet-1K}
\label{app_visual}
\begin{figure}[b]
    \centering
    \begin{subfigure}{0.47\textwidth}
        \centering
        \includegraphics[width=\textwidth]{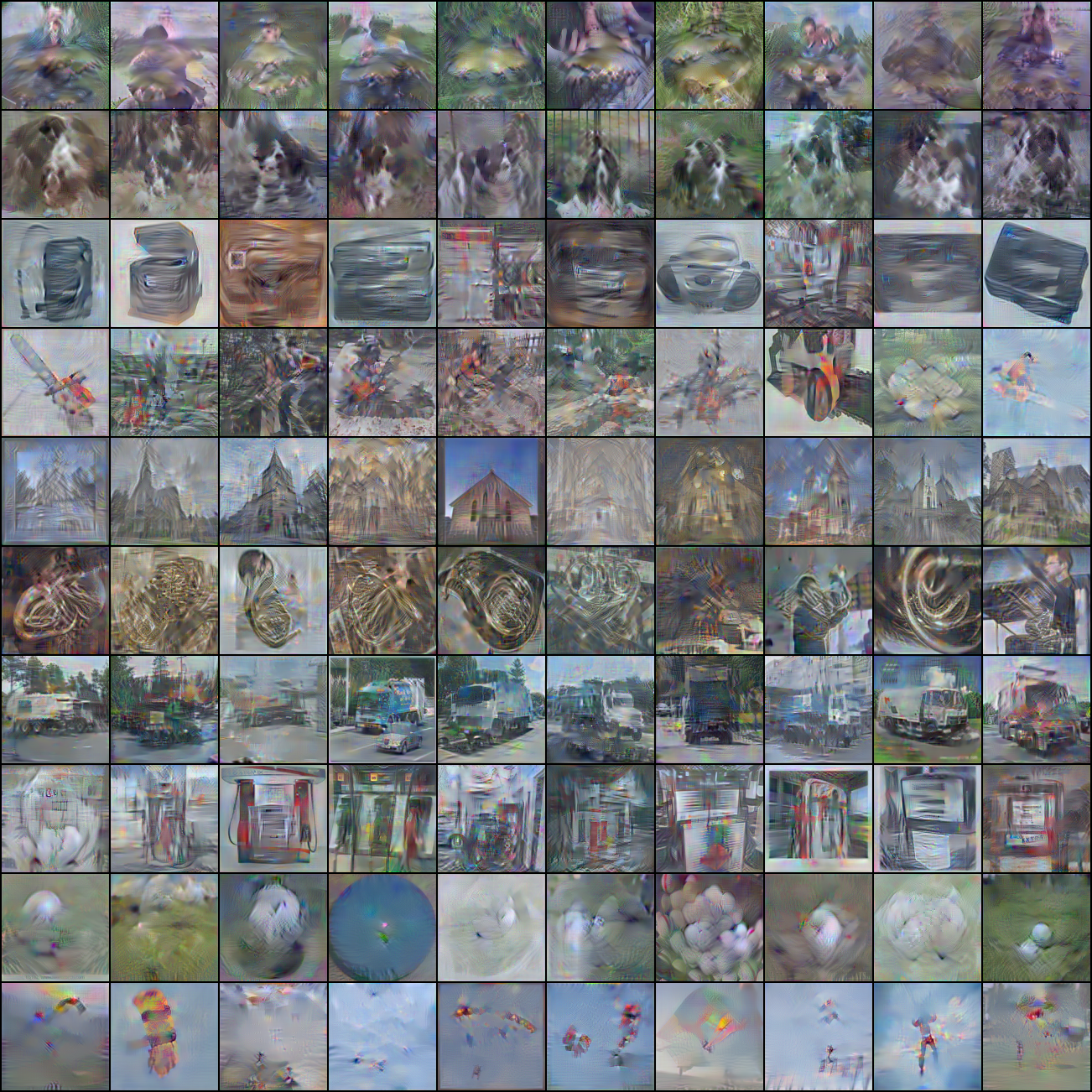}
        \caption{ImageNette}
        \label{}
    \end{subfigure}
    \hfill
    \begin{subfigure}{0.47\textwidth}
        \centering
        \includegraphics[width=\textwidth]{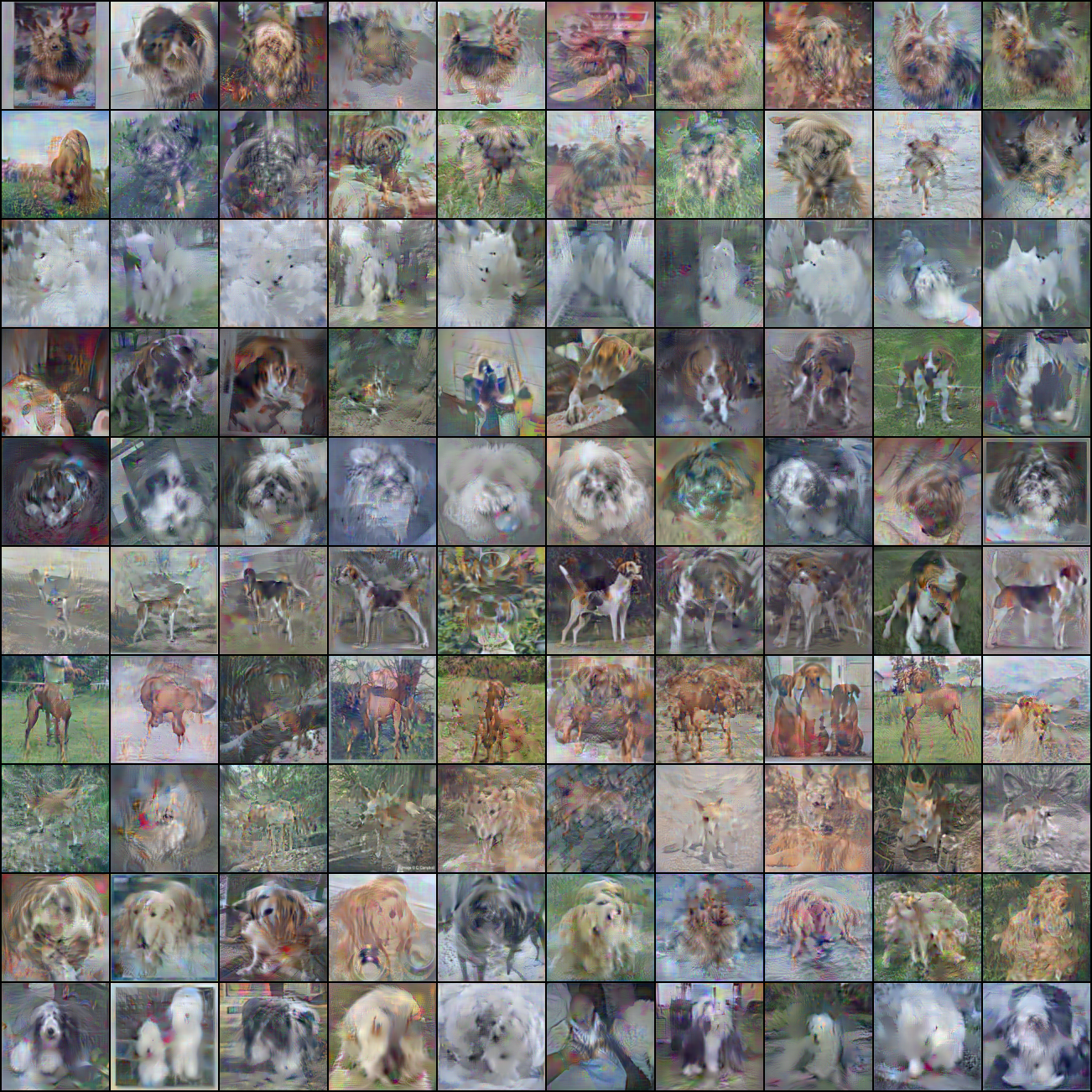}
        \caption{ImageWoof}
        \label{}
    \end{subfigure}
    \caption{}
    \label{fig8}
\end{figure}

In Figure~\ref{fig8} to \ref{fig10}, we present a visualization of distilled images of all ImageNet-1K subsets in Table~\ref{table1}.

\section{More Related Work}
\label{app_rel_work}
In Table~\ref{more_rel_work}, we present a comprehensive summary of previous dataset distillation methods, categorized by different approaches. There are four main categories of dataset distillation: gradient matching, trajectory matching, distribution matching, and generative model-based methods.
Recently, some works~\citep{Yin2023SqueezeRA, Sun2023OnTD, yu2024heavylabelsoutdataset} add knowledge distillation during the evaluation stage of dataset distillation.
~\cite{qin2024labelworththousandimages, loo2024large, chen2024provable, miao2024moreefficienttimeseries, yin2023datasetdistillationlargedata, liu2022dataset, yuan2024colororiented, du2024diversitydrivensynthesisenhancingdataset, xiao2024largescalesoftlabelsnecessary, wei2023sparse, pmlr-v162-vicol22a, izzo2023a, maalouf2023sizeapproximationerrordistilled, lorraine2019optimizingmillionshyperparametersimplicit, yang2024datasetdistillationlearning, cui2022dcbenchdatasetcondensationbenchmark, wu2024ddrobustbenchadversarialrobustnessbenchmark, wu2024visionlanguagedatasetdistillation, xu2024lowranksimilarityminingmultimodal, chen2023comprehensivestudydatasetdistillation}. 

\begin{figure}[htb]
    \centering
    \begin{subfigure}{0.47\textwidth}
        \centering
        \includegraphics[width=\textwidth]{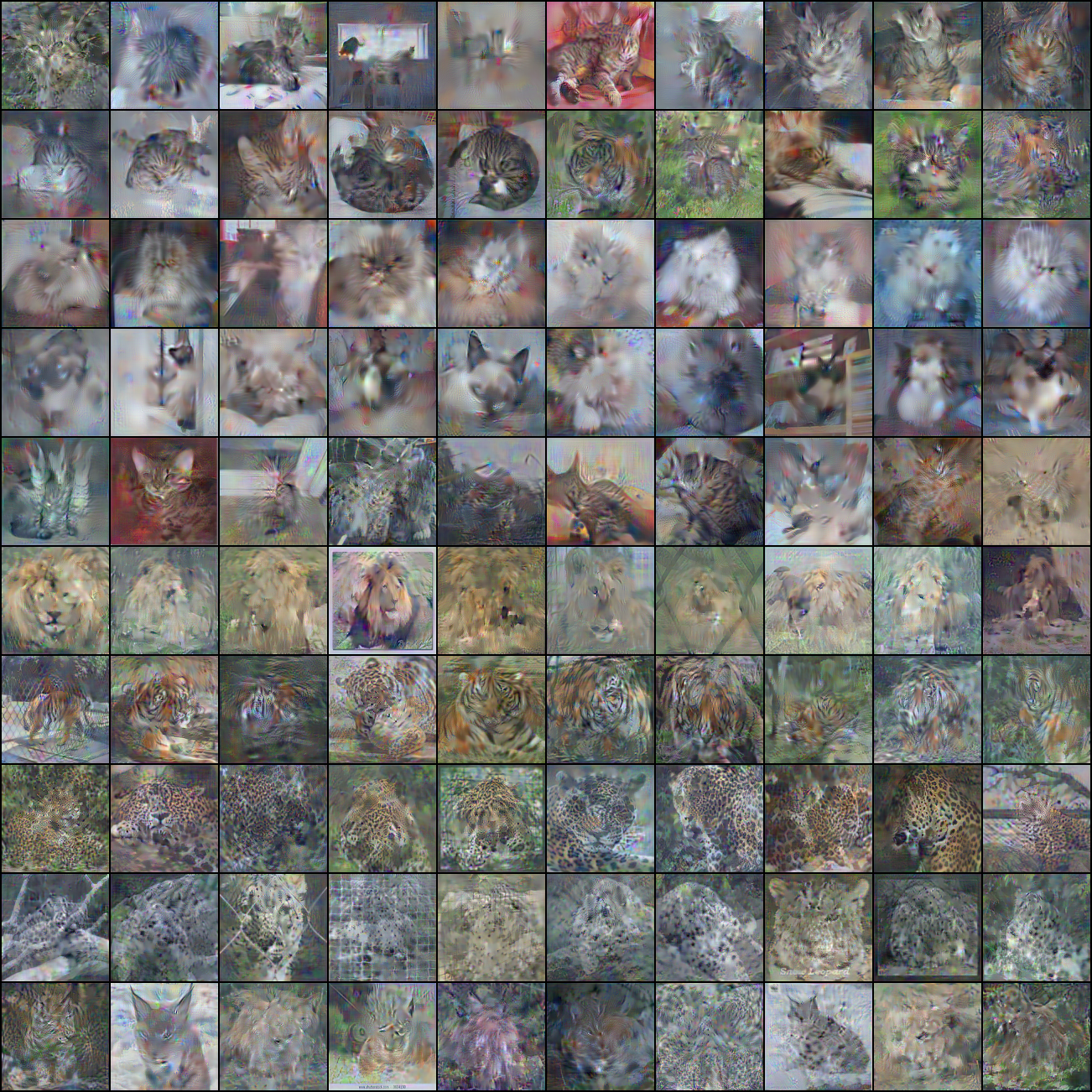}
        \caption{ImageMeow}
        \label{}
    \end{subfigure}
    \hfill
    \begin{subfigure}{0.47\textwidth}
        \centering
        \includegraphics[width=\textwidth]{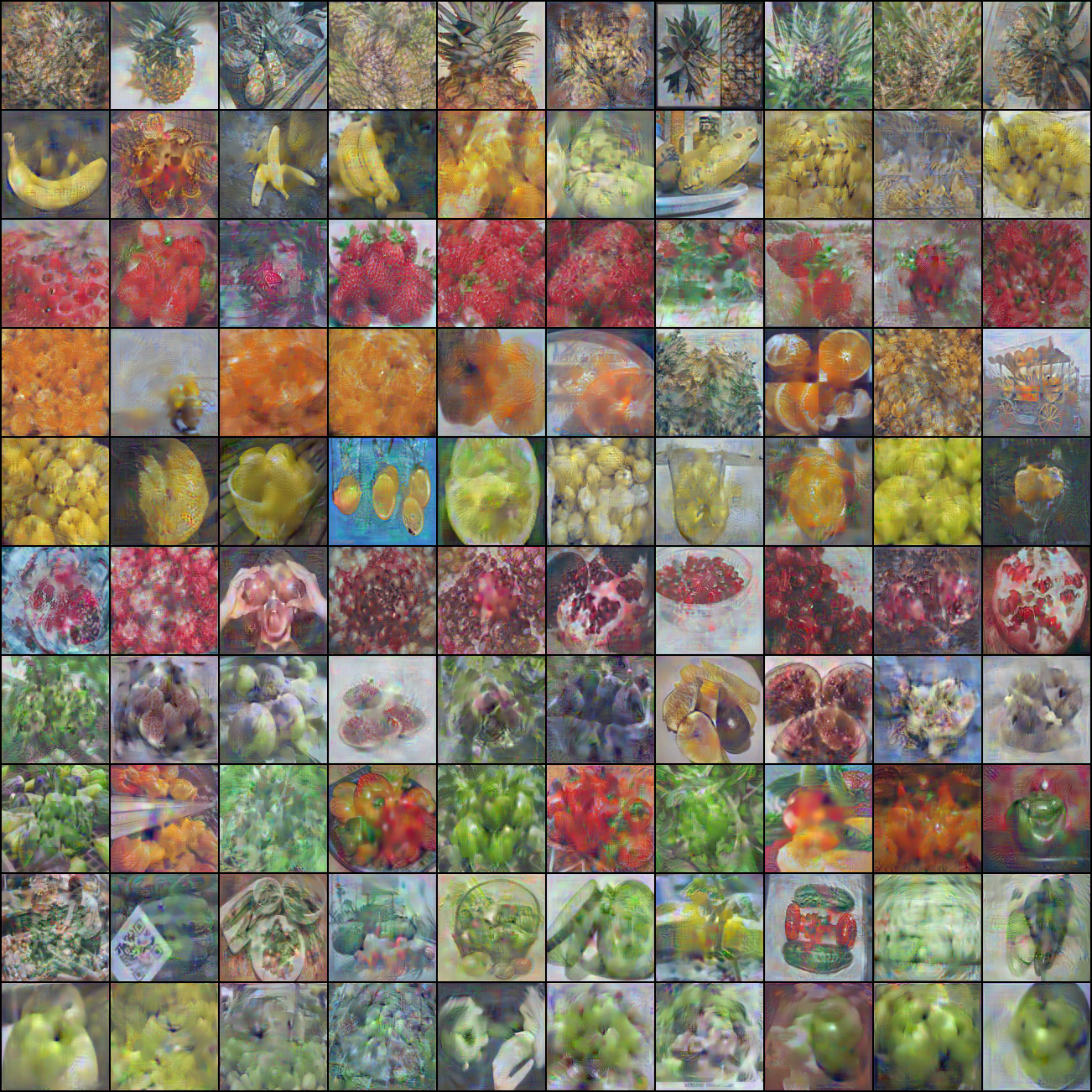}
        \caption{ImageFruit}
        \label{}
    \end{subfigure}
    \caption{}
    \label{fig9}
\end{figure}

\begin{figure}[htb]
    \centering
    \begin{subfigure}{0.47\textwidth}
        \centering
        \includegraphics[width=\textwidth]{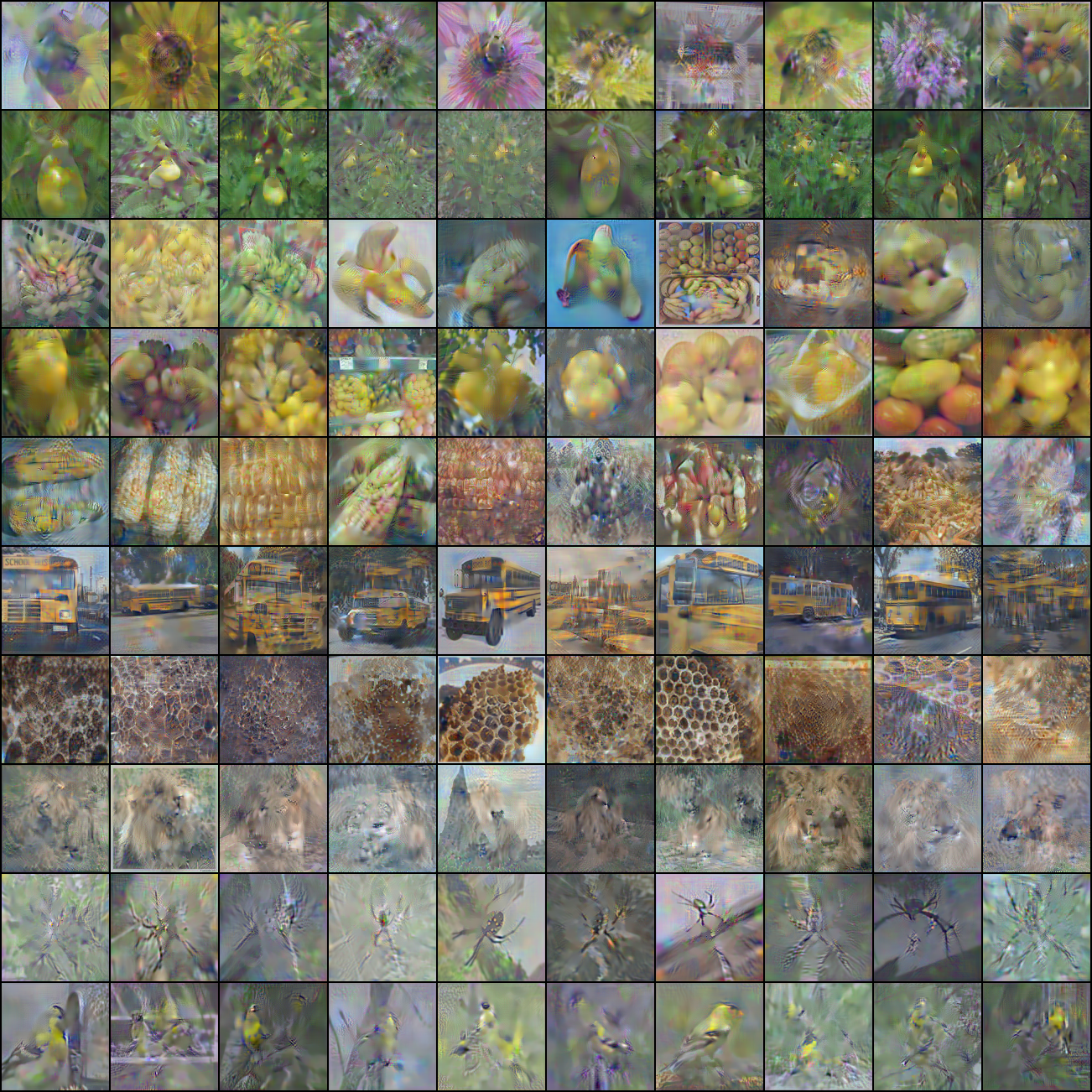}
        \caption{ImageYellow}
        \label{}
    \end{subfigure}
    \hfill
    \begin{subfigure}{0.47\textwidth}
        \centering
        \includegraphics[width=\textwidth]{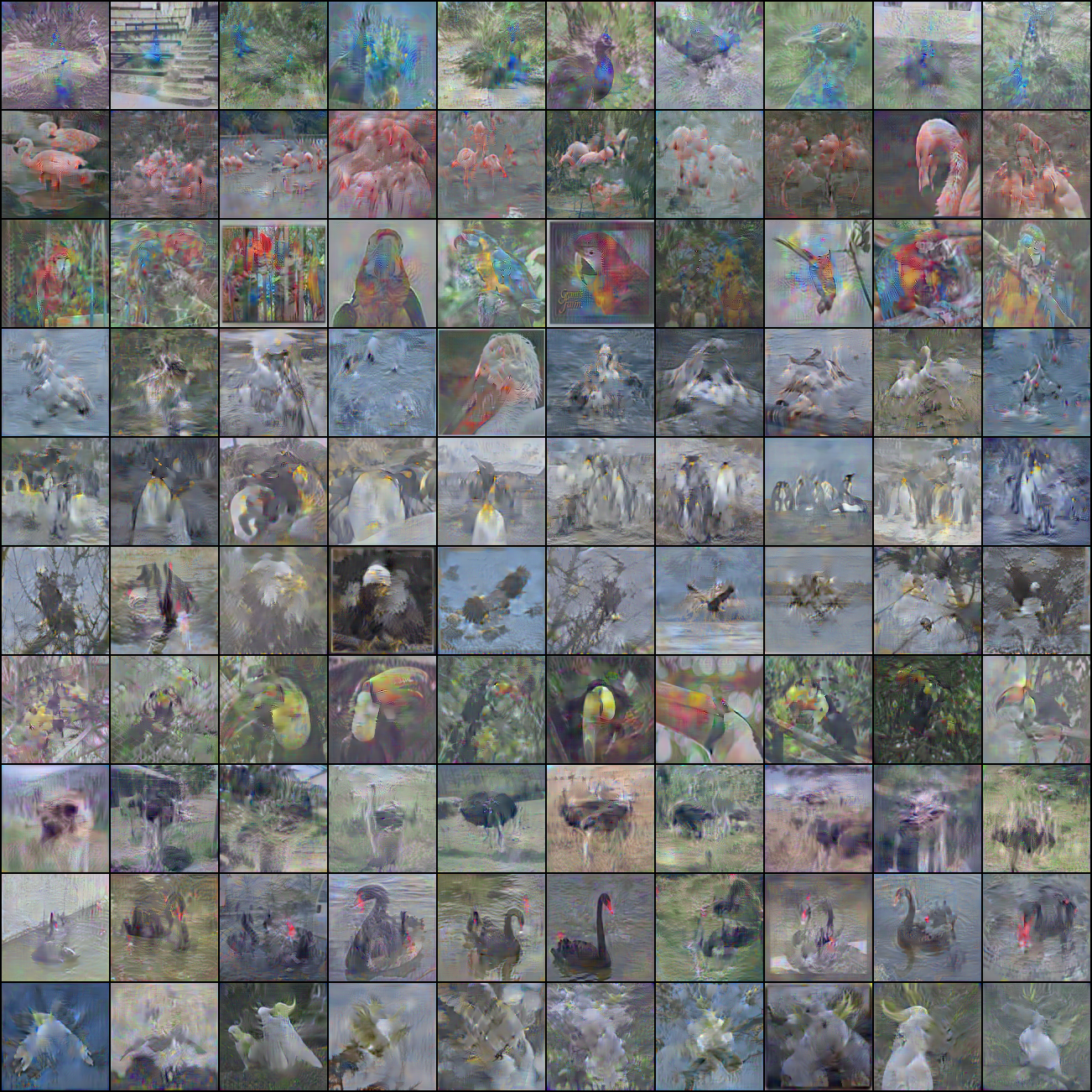}
        \caption{ImageSquawk}
        \label{}
    \end{subfigure}
    \caption{}
    \label{fig10}
\end{figure}

\begin{table}[t]
\centering
\tablestyle{10pt}{1.2}
\begin{tabular}{c|c}
Category  & Method   \\ \shline
\multicolumn{1}{l|}{\multirow{5}{*}{Kernel-based}} & KIP-FC~\citep{KIP-FC}       \\
                                                & KIP-ConvNet~\citep{KIP-ConvNet}      \\
                                                & FRePo~\citep{FRePo}      \\
                                                & RFAD~\citep{RFAD}    \\ 
                                                & RCIG~\citep{RCIG}   \\ \shline

\multicolumn{1}{l|}{\multirow{4}{*}{Gradient-matching}}                          & DC~\citep{DC}       \\
                                                            & DSA~\citep{DSA}      \\
                                                            & DCC~\citep{DCC}      \\
                                                            & LCMat~\citep{LCMat}    \\ \shline
\multicolumn{1}{l|}{\multirow{9}{*}{Trajectory-matching}}                        & MTT~\citep{Cazenavette2022DatasetDB}      \\
                                                            & Tesla~\citep{TESLA}    \\
                                                            & FTD~\citep{Du2022MinimizingTA}     \\
                                                            & SeqMatch~\citep{Du2023SequentialSM} \\
                                                            & DATM~\citep{guo2024lossless}     \\
                                                            & ATT~\citep{ATT}      \\
                                                            & NSD~\citep{NSD}      \\
                                                            & PAD~\citep{li2024prioritizealignmentdatasetdistillation}      \\
                                                            & SelMatch~\citep{lee2024selmatcheffectivelyscalingdataset} \\ \shline
\multicolumn{1}{l|}{\multirow{6}{*}{Distribution-matching}} & DM~\citep{DM}       \\
\multicolumn{1}{l|}{}                                       & CAFE~\citep{Wang2022CAFELT}     \\

\multicolumn{1}{l|}{}                                       & IDM~\citep{Zhao2023ImprovedDM}      \\
\multicolumn{1}{l|}{}                                       & DREAM~\citep{Liu2023DREAMED}    \\
\multicolumn{1}{l|}{}                                       & M3D~\citep{M3D}      \\ 
\multicolumn{1}{l|}{}                                       & NCFD~\citep{wang2025datasetdistillationneuralcharacteristic}      \\ \shline
\multicolumn{1}{l|}{\multirow{7}{*}{Generative model}}      & DiM~\cite{Wang2023DiMDD}       \\
\multicolumn{1}{l|}{}                                       & GLaD~\citep{Cazenavette2023GeneralizingDD}     \\
\multicolumn{1}{l|}{}                                       & H-GLaD~\citep{H-GLaD}      \\
\multicolumn{1}{l|}{}                                       & LD3M~\citep{Moser2024LatentDD}    \\
\multicolumn{1}{l|}{}                                       & IT-GAN~\citep{zhao2022synthesizinginformativetrainingsamples}    \\
\multicolumn{1}{l|}{}                                       & D4M~\cite{D4M}    \\
\multicolumn{1}{l|}{}                                       & Minimax Diffusion~\cite{Gu2023EfficientDD}    \\ \shline
\multicolumn{1}{l|}{\multirow{3}{*}{+ Knowledge distillation for evaluation}} & SRe2L~\citep{Yin2023SqueezeRA}       \\
\multicolumn{1}{l|}{}                                        & RDED~\citep{Sun2023OnTD}     \\
\multicolumn{1}{l|}{}                                        & HeLIO~\citep{yu2024heavylabelsoutdataset}         \\ \shline
\multicolumn{1}{l|}{\multirow{3}{*}{Others}} & MIM4DD~\citep{shang2023mim4ddmutualinformationmaximization}       \\
\multicolumn{1}{l|}{}                                        & DQAS~\citep{zhao2024datasetquantizationactivelearning}     \\
\multicolumn{1}{l|}{}                                        & LDD~\citep{zhao2025distillinglongtaileddatasets}         \\
\end{tabular}
\caption{Summary of previous works on dataset distillation}
\label{more_rel_work}
\end{table}

\end{document}